%% file: 00-main.tex
\documentclass[10pt,twocolumn,letterpaper]{article}

\usepackage{cvpr}              %

\usepackage{graphicx}
\usepackage{amsmath}
\usepackage{amssymb}
\usepackage{booktabs}
\usepackage{algorithm}
\usepackage{algorithmic}
\usepackage{stackrel}
\usepackage{mathtools}
\usepackage{makecell}

\usepackage[dvipsnames]{xcolor}
\usepackage{colortbl}

\definecolor{LightCyan}{rgb}{0.88,1,1}
\DeclareMathOperator*{\argmax}{argmax}

\newcommand\argmaxk{\stackrel{\mathclap{\scriptsize\mbox{k}}}{\argmax}}

\newcommand{\methodName}{\textsc{QB-Norm}\xspace}
\newcommand{\methodNameLong}{Querybank Normalisation\xspace}
\newcommand{\querybank}{querybank\xspace}
\newcommand{\gallery}{\mathcal{G}}
\newcommand{\queryEncoder}{\phi_q}
\newcommand{\galleryEncoder}{\phi_g}

\def\@fnsymbol#1{\ensuremath{\ifcase#1\or *\or \dagger\or \ddagger\or
   \mathsection\or \mathparagraph\or \|\or **\or \dagger\dagger
   \or \ddagger\ddagger \else\@ctrerr\fi}}
\newcommand{\ssymbol}[1]{^{\@fnsymbol{#1}}}

\newcommand\spacebefore{-0.6cm}
\newcommand\spaceafter{-0.25cm}

\makeatletter
\def\blfootnote{\xdef\@thefnmark{}\@footnotetext}
\makeatother

\usepackage[pagebackref,breaklinks,colorlinks]{hyperref}

\usepackage[capitalize]{cleveref}
\crefname{section}{Sec.}{Secs.}
\Crefname{section}{Section}{Sections}
\Crefname{table}{Table}{Tables}
\crefname{table}{Tab.}{Tabs.}

\begin{document}

\title{Cross Modal Retrieval with \methodNameLong}

\author{
Simion-Vlad Bogolin\textsuperscript{1,2,*}
\qquad
Ioana Croitoru\textsuperscript{1,2,*}
\\
Hailin Jin\textsuperscript{3}
\qquad
Yang Liu\textsuperscript{1,4,$\dagger$}
\qquad
Samuel Albanie\textsuperscript{1,5,$\dagger$}
 \\
{\small
\textsuperscript{1}Visual Geometry Group, University of Oxford
\qquad
\textsuperscript{2}Inst. of Mathematics of the Romanian Academy
\qquad
\small
\textsuperscript{3}Adobe Research
} \\
{\small
\textsuperscript{4}Wangxuan Inst. of Computer Technology, Peking University
\qquad
\textsuperscript{5}Department of Engineering, University Cambridge
} \\
}

\maketitle

\begin{abstract}
Profiting from large-scale training datasets, advances in neural architecture design and efficient inference,
joint embeddings have become the dominant approach for tackling cross-modal retrieval.
In this work we first show that, despite their effectiveness, state-of-the-art joint embeddings suffer
significantly from the longstanding ``hubness problem'' in which a small number of gallery embeddings
form the nearest neighbours of many queries.
Drawing inspiration from the NLP literature, we formulate a simple but effective framework
called \methodNameLong (\methodName) that re-normalises query similarities to account for hubs in
the embedding space.
\methodName improves retrieval performance without requiring retraining.
Differently from prior work, we show that \methodName works effectively without concurrent access
to any test set queries. Within the \methodName framework, we also propose a novel similarity
normalisation method, the Dynamic Inverted Softmax, that is significantly more robust than existing approaches.
We showcase \methodName across a range of cross modal retrieval models and benchmarks where it consistently
enhances strong baselines beyond the state of the art.
Code is available at \url{https://vladbogo.github.io/QB-Norm/}.
\end{abstract}

\input{01-intro}
\input{02-related}
\input{03-method}
\input{04-experiments}
\input{05-conclusions}
\input{06-ack}

{\small
\bibliographystyle{ieee_fullname}
\bibliography{refs}
}
\newpage
\input{supp-mat}

\end{document}

%% file: 01-intro.tex
\section{Introduction}
\label{sec:intro}

\blfootnote{
   \textsuperscript{*}Equal contribution.
   \textsuperscript{$\dagger$}Corresponding authors.
}

As the improving price-performance of hardware underpinning
sensors, storage and networking continues to enable the
expansion of humanity's digital archives, the capacity to
efficiently search data takes on greater commercial and scientific importance.
An appealing way to search such data is via \textit{natural language queries},
in which the user describes the target of their search exactly
as they would to another human,
rather than employing specialised database languages such as Structured Query Language (SQL).

\begin{figure}
    \centering
    \includegraphics[trim=8cm 2.5cm 14.4cm 1cm,clip,width=\linewidth]{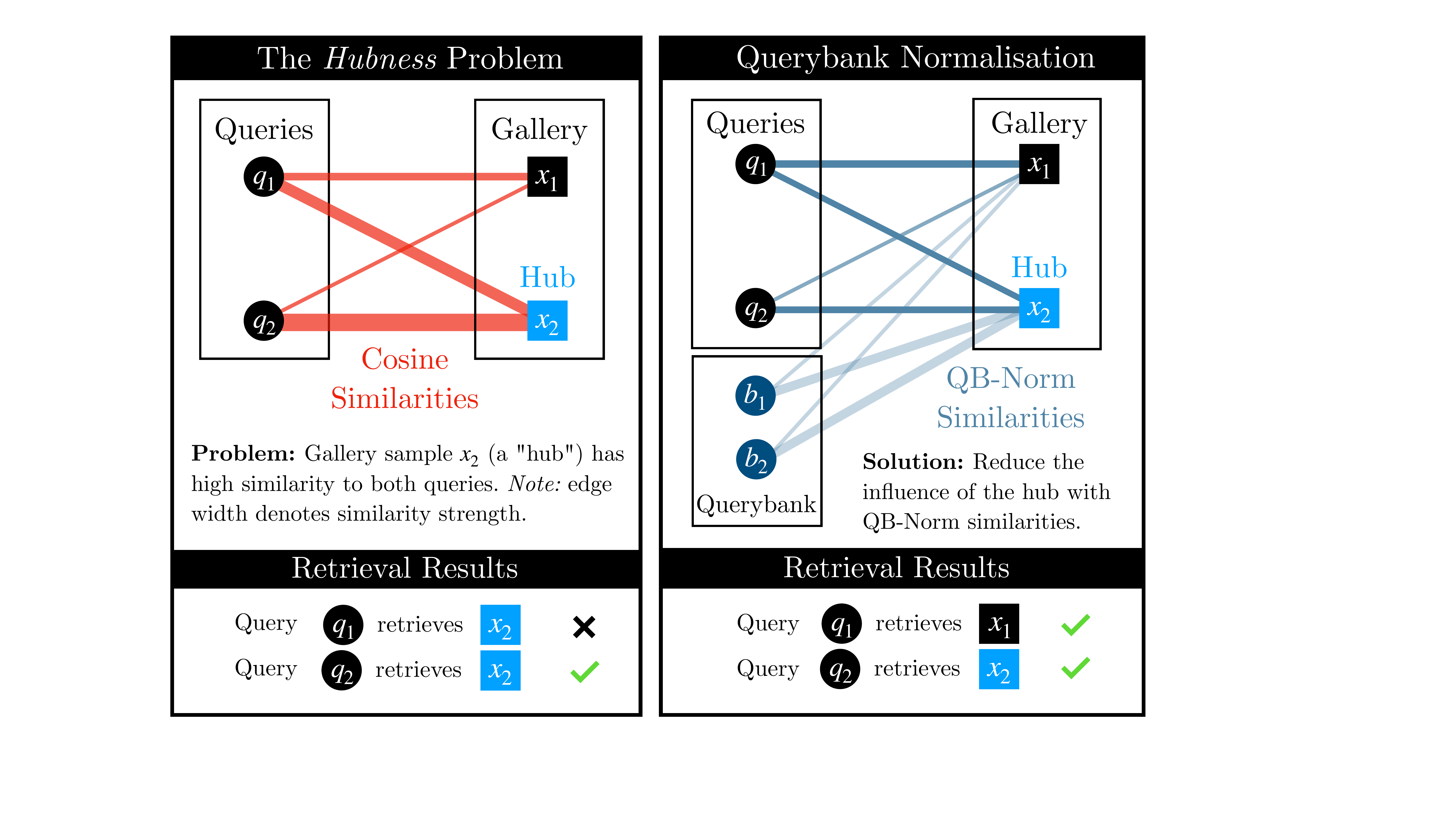}
    \vspace{-0.9cm}
    \caption{
    \textit{Left:} \textbf{The hubness problem.}
    We consider the problem of cross modal retrieval in which 
    queries $q_1$ and $q_2$ are compared
    against a gallery of samples, $x_1$ and $x_2$.
    As we show %
    in Sec.~\ref{sec:method:motivation},
    the high-dimensional joint embeddings
    employed by modern methods
    for cross-modal retrieval suffer from the
    ``hubness problem''~\cite{radovanovic2010hubs}.
    A hub (e.g. $x_2$) is the nearest
    neighbour to multiple queries ($q_1$ and $q_2$),
    producing poor quality retrieval results (bottom left).    
    \textit{Right:} \textbf{\methodNameLong}
    employs a \textit{\querybank} to normalise similarities,
    reducing the similarity of hub $x_2$ to query $q_1$,
    improving the retrieval results (bottom right).
    }
    \mbox{}
    \vspace{-0.7cm}
    \label{fig:teaser}
\end{figure}

Towards this goal, a rich body of research literature has studied
the problem of \textit{cross modal retrieval},
the task of searching a gallery of samples in one modality given
a query in another.
In particular, there has been significant progress
in recent years
for systems that can efficiently search images~\cite{Socher2014GroundedCS}, audio~\cite{oncescu2021audio}
and videos~\cite{Xu2015JointlyMD} with natural language queries by employing
cross modal embeddings.

The dominant cross modal embedding paradigm employs deep neural networks
that project modality-specific samples into a high-dimensional, real-valued
vector space in which they can be directly compared via an appropriate distance metric.
A key challenge for such methods, intrinsic to such high-dimensional spaces,
is the emergence of ``hubs''~\cite{radovanovic2010hubs}---embedding
vectors that appear amongst the nearest neighbour sets of disproportionately
many other embedding vectors (Fig.~\ref{fig:teaser}, left). 
To illustrate this challenge, we show empirically
in Sec.~\ref{sec:method:motivation} and Fig.~\ref{fig:motivation}
that hubness is prevalent among a range of leading retrieval methods.
Hubs have consequences: if left unaddressed,
they lead to a significant degradation in the
search ranking yielded by a retrieval system~\cite{Berenzweig2007thesis}.
The hubness problem has received considerable attention~\cite{Berenzweig2007thesis,radovanovic2010hubs,low2013hubness}
and a number of approaches have been proposed to address
it~\cite{feldbauer2019comprehensive},
with notable contributions in the NLP literature focusing
on bilingual word translation \cite{dinu2014improving,smith2017offline,conneau2018word}.
One contribution of our work is to show how each
of these methods can be interpreted within a single
unifying conceptual framework termed
\methodNameLong (\methodName, Fig.~\ref{fig:teaser}, right),
that employs a \textit{querybank} of samples during
inference to reduce the influence of hubs in the gallery.
We observe that existing methods have two challenges:
(1) To date, these approaches have only been shown to work
with concurrent access to multiple test queries---an assumption that
is impractical for real-world retrieval systems;
(2) They are sensitive to querybank selection,
and indeed actively harm performance for certain querybanks (Tab.~\ref{tab:querybank-domain-gap}). 
To address the first challenge,
we demonstrate through careful experiments
(Tab.~\ref{tab:querybank-train-vs-test}) that
\methodName does \textit{not require}
concurrent access to test queries to be effective.
To address the second challenge,
we propose a new normalisation method,
\textit{Dynamic Inverted Softmax} (DIS),
that operates as a module within the \methodName framework.
We show that DIS provides effective normalisation,
yet is more robust than prior approaches~\cite{dinu2014improving,smith2017offline,conneau2018word}.

We make the following contributions:
(1) We motivate our study by demonstrating that the longstanding
problem of \textit{hubness} remains a significant concern in modern
cross modal embeddings for retrieval;
(2) We propose \methodNameLong (\methodName),
a simple non-parametric framework that brings significant gains
in retrieval performance without requiring model fine-tuning;
(3) We provide the first (to the best of our knowledge)
demonstration that \methodNameLong methods
retain their effectiveness for cross modal retrieval
with no access to test queries beyond the current query;
(4) We propose the Dynamic Inverted Softmax,
a novel normalisation method for \methodNameLong
that is more robust than prior literature;
(5) We show that \methodName is highly effective across a 
broad range of tasks, models and benchmarks.

%% file: 02-related.tex
\section{Related work}
\label{sec:related}

In this section, we summarise prior work from the literature
that relates to our approach, focusing on
\textit{cross-modal retrieval},
\textit{external memory banks}
and
\textit{hubness}.

\noindent \textbf{Cross-modal representations}.
Following initial studies
in psychology~\cite{bruce1986understanding},
early frameworks for cross-modal retrieval
included
Gaussian Mixture Models~\cite{slaney2002semantic} modelling translation via
EM~\cite{duygulu2002object},
Topic Models~\cite{blei2003modeling},
CCA~\cite{rasiwasia2010new},
KCCA\cite{socher2010connecting}
and
rank optimisation~\cite{weston2011wsabie}.
Motivated by the successes of deep metric learning~\cite{chopra2005learning}
and deep visual semantic embeddings~\cite{frome2013devise},
there has since been a Cambrian explosion of cross-modal embedding methods
for text-image retrieval~\cite{kiros2014unifying,Wang2016LearningDS,Karpathy2017DeepVA,Mithun2018WeblySJ,Faghri2018VSEIV},
text-video~\cite{aytar2008utilizing,xu2015jointly,dong2016word2visualvec,mithun2018learning,yu2018joint,wray2019fine,croitoru2021teachtext,bain2021frozen} 
text-audio~\cite{oncescu2021audio},
image-audio~\cite{kidron2005pixels,owens2016ambient,arandjelovic2017look,Nagrani2018LearnablePC,zhao2018sound}
and combinations of all the above~\cite{aytar2017see}.
Recent research spanning these tasks has explored large-scale
pre-training~\cite{miech2019howto100m,radford2021learning},
domain adaptation~\cite{liu2021adaptive,munro2021domain}
and tight integration of multiple sensory modalities into one side
of the embedding space~\cite{miech2018learning,liu2019use,gabeur2020multi}.

\noindent \textit{Similarity search for retrieval: Tricks of the trade.}
A plethora of techniques have been developed to 
support and enhance similarity search for retrieval,
including k-d trees~\cite{Bentley1975MultidimensionalBS},
re-ranking~\cite{Philbin2007ObjectRW,Jgou2008HammingEA},
query expansion~\cite{Chum2007TotalRA,Chum2011TotalRI}, vector compression schemes based on
binary codes~\cite{Gong2011IterativeQA,He2013KMeansHA}
and quantization~\cite{Indyk1998ApproximateNN,Jgou2011ProductQF}
that help address the \textit{curse of dimensionality}~\cite{Page1961AdaptiveCP}.
Algorithms have been developed for approximate
k-nearest neighbour graph construction on CPUs~\cite{Dong2011EfficientKN}
and GPUs\cite{johnson2019billion},
with the latter drawing on product quantization techniques to scale
up to billion-scale searches.

Differently from the work on cross modal representations
and improved similarity search described above,
we focus specifically on tackling the problem
of \textit{hubness} in cross-modal
embeddings, which we demonstrate (Sec.~\ref{sec:method:motivation})
to be a widespread issue among
leading cross-modal embedding frameworks.

\noindent \textbf{Memory bank augmented architectures.}
Memory banks in various forms have been studied as useful extensions to neural network architectures
to facilitate general
problem-solving~\cite{Graves2014NeuralTM,zaremba2015reinforcement,greve2016evolving,santoro2016meta},
better image captioning~\cite{Park2017AttendTY,xu2020interactive,cornia2020meshed}
and summarisation~\cite{Kim2019AbstractiveSO,Lee2018AMN},
enhance self-supervised
training dynamics~\cite{He2020MomentumCF, bulat2021improving,liu2018unsupervised}
and to provide a mechanism to deal with rare instances~\cite{Kaiser2017LearningTR,Yoo2019ColoringWL}.
Our proposed \methodNameLong framework likewise stores embedding samples in an external memory bank,
but targets a very different problem to these works, namely hubness mitigation.

\noindent \textbf{The Hubness Problem.}
The \textit{hubness problem} was formally characterised
by Radovanovic et al.~\cite{radovanovic2010hubs},
who observed that in points sampled from a distribution 
with high intrinsic dimensionality, the distribution of ``k-occurrences''
(the number of times a point appears in the k nearest neighbours of other points)
skews heavily to the right.
Although there is disagreement about the cause of hubness~\cite{low2013hubness}, it has been conceptually
linked~\cite{Berenzweig2007thesis} to \textit{distance concentration} in high-dimensions
(high-dimensional points lie close to a hypersphere centred on the data mean, i.e., they all exhibit a
similar distance to the mean~\cite{franccois2007concentration}).
It is thought that hubs then result from this phenomenon through the non-negligible variance in the
distribution of distances to the mean in finite dimensions~\cite{radovanovic2010hubs}.

\begin{figure*}[!ht]
    \centering
    \includegraphics[width=\linewidth]{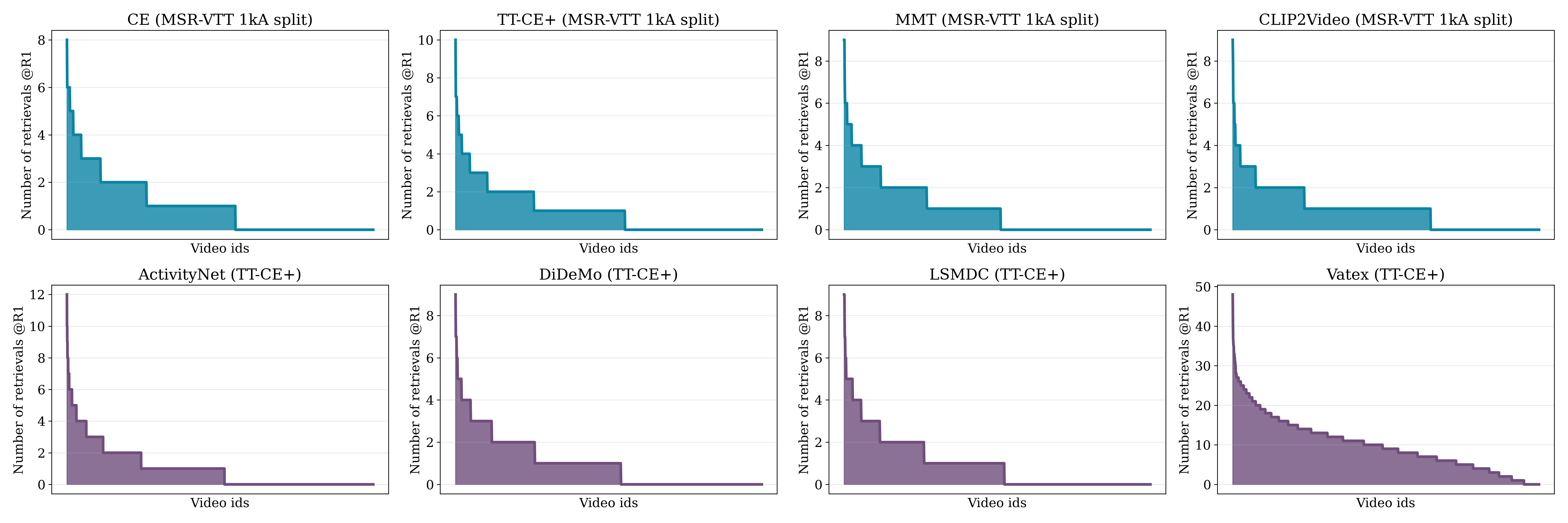}%
    \vspace{-0.4cm}
    \caption{
    \textbf{Hubness is pervasive in leading methods for text-video retrieval}.
    The charts depict the distribution of the number of times each gallery video was
    retrieved by test set queries (x-axes video ids are ordered by decreasing retrieval count).
    \textit{Top row (different models):} 
    We report retrieval distributions for
    CE~\cite{liu2019use},
    TT-CE+~\cite{croitoru2021teachtext},
    MMT~\cite{gabeur2020multi} and
    CLIP2Video~\cite{fang2021clip2video}
    on the MSR-VTT benchmark~\cite{xu2016msr}.
    \textit{Bottom row (different datasets):} 
    We report retrieval distributions for the TT-CE+~\cite{croitoru2021teachtext}
    method on four additional datasets,
    DiDeMo~\cite{Hendricks2017LocalizingMI},
    LSMDC~\cite{Rohrbach2015ADF},
    VaTeX~\cite{Wang2019VaTeXAL},
    and
    ActivityNet-captions~\cite{Krishna2017DenseCaptioningEI},
    In all instances, we observe strong \textit{hubness}, in which a small number of videos are
    retrieved disproportionately often, damaging performance.
    }
    \mbox{}\vspace{-0.7cm} \\
    \label{fig:motivation}
\end{figure*}

\noindent \textit{Hubness Mitigation.}
One paradigm has focused on \textit{rescaling} the similarity space to account for asymmetries
in nearest neighbour relations~\cite{schnitzer2012local}---a process that can be achieved through
both local~\cite{ZelnikManor2004SelfTuningSC,jegou2007contextual} and global~\cite{schnitzer2012local}
scaling schemes.
Another work has focused on addressing the hub-like tendency of centroids in the data
through Laplacian-based kernels~\cite{Suzuki2012InvestigatingTE} and
centring~\cite{suzuki2013centering,Hara2015LocalizedCR}.
Fedbauer et al. provide a comprehensive empirical comparison of these families of methods~\cite{feldbauer2019comprehensive} and note that while effective, these
approaches scale quadratically, making their naive application unsuitable for large
datasets.
One exception is the CENT method~\cite{suzuki2013centering}, however we did
not find this approach to be effective (experiments are provided in the supplementary).
In the zero-shot learning literature, works have sought to address hubness by mapping (text)
targets back into the (image) query space~\cite{shigeto2015ridge,zhang2017learning}, and by minimising proxies
for hubness~\cite{Liu2020HALIT} and skewness in the k-occurrences distribution~\cite{Cheraghian2019MitigatingTH}
to improve 3D few-shot learning performance.
More closely related to our work,~\cite{dinu2014improving} propose general
retrieval schemes for which queries are matched with targets for which they form the nearest neighbour.
This work was built upon the NLP literature by~\cite{conneau2018word}, who propose a cross-domain local
scaling scheme (which can be integrated into the loss~\cite{Joulin2018LossIT}),
and by~\cite{smith2017offline}, who introduce the \textit{Inverted Softmax} (IS)
to mitigate hubness when translating between dictionaries in different languages.
We discuss the relationship of our approach to~\cite{dinu2014improving,conneau2018word,smith2017offline}
in more detail in Sec.~\ref{sec:method} and compare these methods with our proposed
\textit{Dynamic Inverted Softmax} in Sec.~\ref{sec:experiments}.

Also related to our work, \cite{liu2019strong,chenteam,Zhao2020RUC} enforce a bipartite matching
constraint between queries and test set items by applying an IS over the full set of test
queries---a constraint that is unrealistic for practical retrieval systems that experience continuous
operation from users. 
One contribution of this paper is to demonstrate that concurrent access to test queries is \textit{not required}.
A second contribution of our work, not considered in prior work, is to show that the techniques proposed
above can actively damage retrieval performance for particular querybank selections, an issue that we address
with our proposed Dynamic Inverted Softmax.

%% file: 03-method.tex
\section{Method}
\label{sec:method}

We first define the task of retrieval with
cross modal embeddings (Sec.~\ref{sec:method:task}),
before outlining the motivation for our work by
examining the hubness problem in the context of
text-video retrieval (Sec.~\ref{sec:method:motivation}).
Next, we introduce the \methodNameLong framework
(Sec.~\ref{sec:method:framework}),
generalising several existing approaches to address this issue.
Finally, we explore designs for framework components
and introduce the proposed Dynamic Inverted Softmax
for robust similarity normalisation
(Sec.~\ref{sec:method:framework-components}).

\subsection{Task definition}
\label{sec:method:task}

Given a \textit{gallery}, $\gallery$, of samples in one modality, $m_g$
and a query, $q$, in another modality, $m_q$, the objective of cross modal
retrieval is to rank the gallery samples according to how 
well they match the query.
We study this problem within the framework of learning
\textit{cross modal embeddings}~\cite{frome2013devise}:
specifically, we seek to learn a pair of encoders,
$\queryEncoder$ and $\galleryEncoder$,
that map each query, $q$, and gallery sample, $g$, into a shared
real-embedding space, $\mathbb{R}^C$,
such that $\queryEncoder(q)$ and $\galleryEncoder(g)$ are close if and only
if $q$ is similar to $g$.
We assume that we are given access to a training set of $T$
corresponding query and gallery samples
$\{(q_i, g_i)\}_{i=1}^T$ for the purposes of learning the embeddings.
However, the queries and gallery used to evaluate
retrieval performance (i.e. the test set) are unseen during training.

The choice of similarity measure used to define a ``good match''
is determined by the application domain.
For instance,
in the task of text-video retrieval with natural language queries,
the objective is to rank a gallery of videos according to how well
their content is described by a written free-form text
query~\cite{miech2018learning},
whereas in image-audio retrieval the objective is typically
to obtain audio samples from the gallery that share the same
semantic category as the image query~\cite{Arandjelovi2018ObjectsTS}.
In this work, we focus particularly on cross modal retrieval tasks
with natural language queries, for two reasons:
(1) these tasks have received limited attention in the
hubness mitigation literature,
(2) hubness has been shown to be particularly prevalent in
embeddings with high \textit{intrinsic dimensionality}~\cite{radovanovic2010hubs}.
Since natural language queries can express more complex
concepts than individual words (such as those considered
in zero-shot learning image labelling tasks~\cite{dinu2014improving}),
we expect might expect natural language queries to naturally induce
cross modal embeddings with greater intrinsic dimensionality,
and thus may have greater potential to benefit from hubness mitigation.

\subsection{Motivation}
\label{sec:method:motivation}

It has long been observed that
high-dimensional embedding spaces are prone to
\textit{hubness}~\cite{radovanovic2010hubs},
in which a small proportion of samples appear disproportionately
frequently among the set of k-nearest neighbours of all embeddings.
As noted by Berenzweig~\cite{Berenzweig2007thesis}, this property
can have damaging consequences for retrieval systems that employ
nearest neighbour search to find the best gallery match for a given
query.
To illustrate this issue, we consider the problem of video
retrieval with natural language queries.
We plot the distribution of the number of
times each gallery video was retrieved on the
MSR-VTT retrieval benchmark~\cite{xu2016msr} 
for an array of text-video retrieval methods,
including
CE~\cite{liu2019use},
TT-CE+~\cite{croitoru2021teachtext},
MMT~\cite{gabeur2020multi}
and CLIP2Video~\cite{fang2021clip2video},
the latter of which represents the current
state of the art on this benchmark.
In each case, we see striking evidence of hubness---a small number of videos
are retrieved extremely often, while others are not retrieved at all.
This phenomenon is not limited to a particular retrieval model,
suggesting that the issue is not readily addressed by the use of
multiple video modalities,
attention mechanisms and
large-scale pretraining 
implemented in various combinations by these approaches.

\subsection{\methodNameLong}
\label{sec:method:framework}

To address the hubness issues observed among cross modal embeddings for
text-video retrieval in the previous section, we first turn to the existing
literature on hubness mitigation.
As noted in Sec.~\ref{sec:related}, hubness effects have been studied in
several problem domains, including
Zero-Shot Learning~\cite{dinu2014improving,shigeto2015ridge},
NLP~\cite{smith2017offline,conneau2018word},
biomedical statistics~\cite{schnitzer2012local}
and music retrieval~\cite{schnitzer2012local}.
Among this literature, we are particularly interested in methods
that can be applied in a practical cross modal retrieval setting,
namely, those methods whose complexity scales at most linearly with the
size of the gallery (rather than quadratic complexity methods that seek
to address hubness within a fixed embedding space~\cite{feldbauer2019comprehensive}).
To clarify relationships between existing approaches,
we cast them into the \methodNameLong framework (Fig.~\ref{fig:teaser}),
which comprises two components,
\textit{querybank construction} and
\textit{similarity normalisation},
described next:

\noindent \textbf{Querybank construction.}
To mitigate hubness in the cross modal embedding space,
we seek to alter the similarities between embeddings in
a way that \textit{minimises the influence of hubs}.
To adjust similarities, we first construct a \textit{querybank} of $N$
samples, $\mathcal{B} = \{b_1, \dots, b_N\}$ from the
query modality, $m_q$, which will serve
as a \textit{probe} to measure the hubness of gallery samples.

\noindent \textbf{Similarity normalisation.}
To normalise similarities to account for hubs, we assume access to a query, $q$,
trained encoders $\queryEncoder$ and $\galleryEncoder$,
querybank $\{b_1, \dots, b_N\}$,
and a gallery $\gallery$.
For each $g_j \in \gallery$, we first compute a \textit{probe vector},
$p_j \in \mathbb{R}^N$,
$p_j(i) = \text{sim}(\queryEncoder(b_i), \galleryEncoder(g_j))$
where $\text{sim}(\cdot, \cdot)$ denotes similarities
in the cross modal embedding
space (e.g. cosine similarity).
The probe vectors are then stacked to form a probe matrix $P \in \mathbb{R}^{|\gallery| \times N}$.
Similarly, we compute for each query a vector of
\textit{unnormalised similarities},
$s_q \in \mathbb{R}^{|\gallery|}$,
$s_q(j) = \text{sim}(\queryEncoder(q), \galleryEncoder(g_j))$.
Here $j\in \{1, \dots, |\gallery|\}$ indexes over all gallery elements.
Finally, we define a querybank normalisation function,
$\methodName: \mathbb{R}^{|\gallery|} \times \mathbb{R}^{|\gallery| \times N}   \to \mathbb{R}^{|\gallery|}$,
which yields, for each query $q$ and gallery $\gallery$,
a vector of querybank normalised similarities,
$\eta_q = \methodName(s_q, P) \in \mathbb{R}^{|\gallery|}$.
Various candidates for $\methodName(\cdot)$ are discussed
in Sec.~\ref{sec:method:framework-components}.

In practice, the probe matrix employed for similarity normalisation can be
precomputed and re-used across all queries
(improving computational efficiency at the cost of higher memory).
An overview of the resulting \methodName algorithm,
and its application to ranking gallery samples for a
collection of queries, $\mathcal{Q}$,
is summarised in Alg.~\ref{alg:querybank-normalisation}.

\begin{algorithm}
\caption{Ranking with \methodNameLong}
\label{alg:querybank-normalisation}
\begin{algorithmic}[1]
\REQUIRE queries, $\mathcal{Q} \subset m_q$
\REQUIRE gallery, $\gallery \subset m_g$
\STATE \textbf{Querybank construction.}
\STATE Construct querybank,  $\mathcal{B} = \{b_1, \dots, b_N \} \subset m_q$
\STATE \textbf{Similarity normalisation: } 
\STATE \textit{Precompute querybank probe matrix}
\begin{ALC@g}
\FOR  {gallery sample $g_j \in \mathcal{G}$}
\FOR  {querybank sample $b_i \in \mathcal{B}$}
\STATE Compute probe matrix entry $P(j, i) = \text{sim}(\queryEncoder(b_i), \galleryEncoder(g_j)) \in \mathbb{R}$
\ENDFOR
\ENDFOR
\end{ALC@g}
\STATE \textit{query computations: \methodName similarities}
\begin{ALC@g}
\FOR  { query $q \in \mathcal{Q}$}
\FOR  { gallery sample $g_j \in \gallery$}
\STATE Compute unnormalised similarity
$s_q(j) = \text{sim}(\queryEncoder(q), \galleryEncoder(g_j))$
\ENDFOR
\STATE $\eta_{q} = \methodName(s_q, P) \in \mathbb{R}^{|\gallery|}$.
\STATE search ranking = argsort($\eta_q$)
\ENDFOR
\end{ALC@g}
\end{algorithmic}
\end{algorithm}

\subsection{Design choices}
\label{sec:method:framework-components}

The \methodNameLong framework admits a number of viable choices
for both querybank construction and similarity normalisation.
To illustrate this point, we first cast three techniques for hubness
mitigation proposed in the NLP literature into the framework.
We then introduce our proposed alternative,
the Dynamic Inverted Softmax.

\noindent \textit{Globally-Corrected (GC) retrieval}~\cite{dinu2014improving}.
This approach, originally introduced for the tasks of bilingual translation and
zero-shot learning, can be implemented by constructing the
querybank from the full set of test queries, $\mathcal{Q}$,
(or all semantic labels, in the cross modal setting of zero-shot image labelling).
For their bilingual translation task, the authors
supplement their querybank by an additional randomly
sampled collection of instances from $m_q$, which improved performance.
The normalised similarity corresponding to $q$
and gallery vector $g_j$ is defined via
$\eta_q(j) = -(\text{Rank}(s_q(j), p_j) - s_q(j)) \in \mathbb{R}$,
where
$\text{Rank}: \mathbb{R} \times \mathbb{R}^N \rightarrow \{0, \dots, N\}$
returns the rank of the first argument
with respect to the array of elements
in the second argument.

\noindent \textit{Cross-Domain Similarity Local Scaling (CSLS)}~\cite{conneau2018word}.
Introduced for the task of bilingual word translation,
CSLS constructs an initial querybank comprising all
possible queries (corresponding to source vocabulary samples),
then employs a different subset of the querybank to normalise
each gallery sample.
Let $\hat{p}_j \in \mathbb{R}^K$ denote the probe vector, $p_j$, restricted
to the $K$ querybank samples that are most similar to gallery sample $g_j$.
Similarly, let $\hat{s}_q \in \mathbb{R}^K$ denote the
unnormalised similarity vector, $s_q$, restricted to the $K$ gallery samples
that are most similar to query $q$. Then the normalised similarity
is computed via:
$\eta_q(j) = 2 s_q(j) - \frac{1}{K}\mathbf{1}^T \hat{s}_q - \frac{1}{K}\mathbf{1}^T \hat{p}_j \in \mathbb{R}$.

\noindent \textit{Inverted Softmax (IS)}~\cite{smith2017offline}.
Targeting bilingual word translation,
this method constructs a querybank from the source vocabulary
(corresponding to all possible queries of interest).
For practical implementations,
the authors recommend to uniformly randomly subsample a feasible number of queries.
\textit{Similarity normalisation} is implemented via:
\vspace{-0.1cm}
\begin{equation}
\eta_q(j) =
\frac{\exp(\beta \cdot s_q(j)) }{\mathbf{1}^T\exp[\beta \cdot p_j]} \in \mathbb{R}
\label{eqn:inverted-softmax}
\vspace{-0.1cm}
\end{equation}
where $\exp[\cdot]$ denotes elementwise exponentiation and $\beta$
 is a hyperparameter referred to as the ``inverse temperature''.

\noindent \textit{Dynamic Inverted Softmax} (DIS).
In experiments with the methods described above
(discussed in detail in Sec.~\ref{sec:experiments})
we observed an important practical issue:
if the querybank does not effectively cover the space
containing the gallery, performance is severely degraded
such that it falls below the performance of unnormalised similarities.
This characteristic renders them less desirable for a
general-purpose solution: we would like something that
not only enhances performance in favourable conditions,
but also ``does no harm'' when curating a querybank
to match the gallery closely is challenging.
To address this issue,
in addition to the querybank probe matrix described in Alg.~\ref{alg:querybank-normalisation},
we also precompute a \textit{gallery activation set},
$\mathcal{A} =
\{j : j \in \argmaxk_l s(b_i, g_l), i \in \{1, \dots, N\} \}$.
Here, the notation $\argmaxk_l f(l)$
denotes the $k$-max-select operator
that returns the $k$ values of $l$ that maximise $f(l)$
(like $j$, $l$ also runs over the gallery indices and $k$ is set as a hyperparameter).
Intuitively, this set contains the indices of gallery vectors
that our querybank probe has identified as potential hubs.
We create a Dynamic Inverted Softmax by activating
the inverted softmax only for nearest neighbour retrievals that
fall within this set:
\vspace{-0.1cm}
\begin{equation}
\eta_q(j) =
\begin{cases}
\frac{\exp(\beta \cdot s_q(j)) }{\mathbf{1}^T\exp[\beta \cdot p_j]} & \text{if } \argmax_l s_q(l) \in \mathcal{A} \\
s_q(j) & \text{otherwise}
\end{cases}
\label{eqn:dynamic-inverted-softmax}
\vspace{-0.1cm}
\end{equation}

Since $s_q(j)$ is computed as an intermediate step in Eqn.~\ref{eqn:inverted-softmax}, the only additional cost
incurred by the Dynamic Inverted Softmax over the standard
Inverted Softmax stems from the argmax operation in Eqn.~\ref{eqn:dynamic-inverted-softmax}.
Fortunately, this computation can be performed extremely efficiently
with almost no loss in precision, even at the scales of billions
of gallery samples~\cite{johnson2019billion}.
We show through experiments in Sec.~\ref{sec:experiments},
the Dynamic Inverted Softmax is significantly more robust than
GC, CSLS and IS: crucially, it does not harm performance
when employed with suboptimal querybank selection.

%% file: 04-experiments.tex
\section{Experiments}
\label{sec:experiments}

In this section,
we first briefly describe the datasets and metrics
used for our experiments (Sec.~\ref{sec:experiments:datasets}).
We then conduct a series of experiments that:
(i) demonstrate our claim
that \methodName is effective without concurrent
access to more than one test query;
(ii) investigate the influence of querybank size;
(iii) compare the \textit{Dynamic Inverted Softmax}
against prior methods;
(iv) ablate other \methodName components
(Sec.~\ref{sec:experiments:querybank-normalisation}).
Finally, we demonstrate the generality of \methodNameLong
by applying it to a broad range of models,
tasks and datasets (Sec.~\ref{sec:experiments:benchmarks}).

\subsection{Datasets and Evaluation Metrics}
\label{sec:experiments:datasets}

We conduct experiments on standard benchmarks
for text-video retrieval:
MSR-VTT~\cite{xu2016msr},
MSVD~\cite{Chen2011CollectingHP},
DiDeMo~\cite{Hendricks2017LocalizingMI},
LSMDC~\cite{Rohrbach2015ADF},
VaTeX~\cite{Wang2019VaTeXAL} and
QueryYD~\cite{oncescu20queryd}.
We also investigate \methodName on text-image retrieval
(MSCoCo~\cite{Chen2015MicrosoftCC}),
text-audio retrieval (AudioCaps~\cite{Kim2019AudioCapsGC}),
and image-to-image retrieval
(CUB-200-2011~\cite{Wah2011TheCB}, Stanford Online Products~\cite{Song2016DeepML}).
Detailed descriptions of each
dataset are deferred to the supplementary.
We report standard retrieval performance metrics:
R@K (recall at rank K, higher is better) and
MdR (median rank, lower is better).
For each study, we report the mean and standard
deviation over three randomly seeded runs.

\subsection{\methodNameLong}
\label{sec:experiments:querybank-normalisation}

We conduct initial studies on the MSR-VTT
benchmark for text-video retrieval using
TT-CE+~\cite{croitoru2021teachtext} to address
a series of questions relating to \methodNameLong.

\noindent \textbf{Do we need access to more than
one test query at a time to mitigate hubness?}
Prior work has investigated the use of IS for image
and video retrieval with natural language queries,
but only by assuming simultaneous access to the
full test set of queries to construct the querybank~\cite{liu2019strong,chenteam,Zhao2020RUC}.
The motivation for this approach~\cite{liu2019strong} is to enforce
a bipartite matching constraint that encodes the prior knowledge
that each test query maps to exactly one gallery sample.
Unfortunately, this approach is impractical to deploy for
real world systems that experience sequential user queries.
Therefore, we first ask whether we require concurrent access
to all test set queries by constructing an alternative querybank
from the training set.
We evaluate performance with \methodName using DIS normalisation
in which we construct querybanks from:
(i) all test set queries;
(ii) all validation set queries:
(iii) a randomly subsampled subset of the training set matching
the size of the test set
(resampled once for each trained model to estimate variance).
The results are reported in Tab.~\ref{tab:querybank-train-vs-test}.
Remarkably, we observe that 
\textit{training set querybanks perform comparably to test set
querybanks}.
Given this finding, we conclude that
\textit{test set querybanks are not necessary to mitigate hubness}.
We therefore restrict all querybank construction to
use training set samples for all remaining experiments,
ensuring valid comparisons on standard retrieval benchmarks.

\input{tables/querybank-train-vs-test}

\begin{figure}
    \centering
    \includegraphics[trim={0cm 0.5cm 0cm 0.2cm},clip,width=\linewidth]{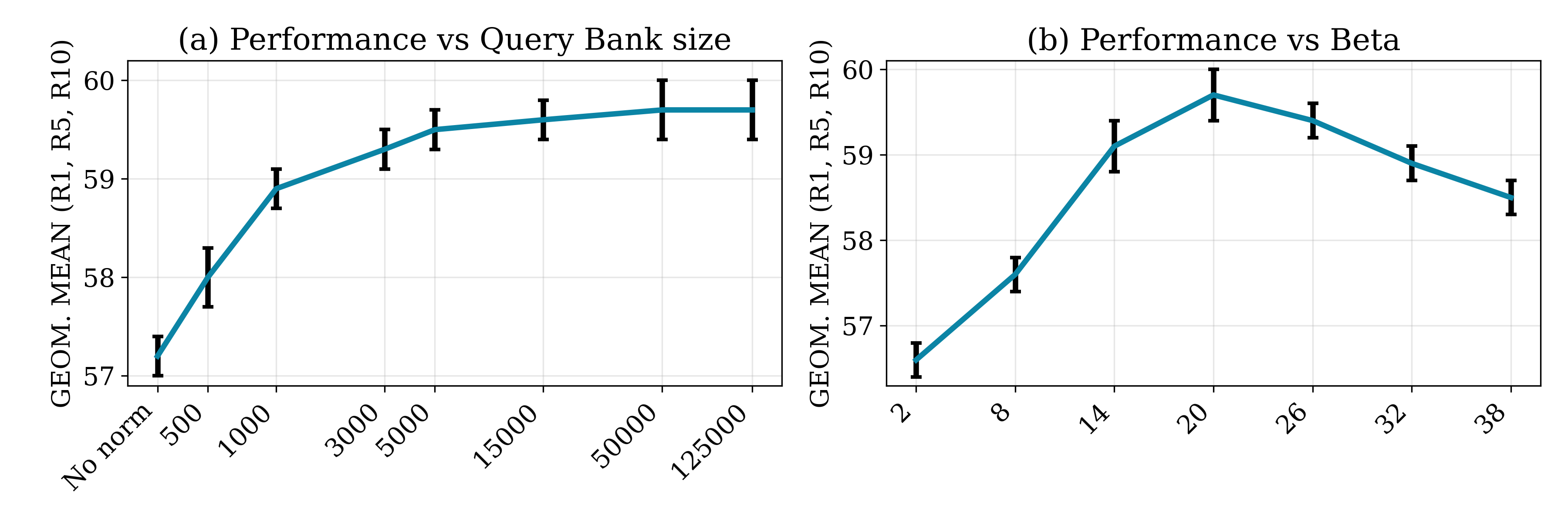}
    \vspace{-0.3cm}
    \caption{Retrieval results reported for a
    TT-CE+~\cite{croitoru2021teachtext} model
    on the MSR-VTT~\cite{xu2016msr} benchmark
    for text-video retrieval with \methodName \textit{DIS} normalisation.
    \textit{Left:} \textit{The influence of querybank size on retrieval performance on the \texttt{validation} split of MSR-VTT.}
    We observe that performance grows steadily with increasing
    querybank size, but saturates.
    \textit{Right:}
    \textit{The influence of inverse temperature, $\beta$, on the \texttt{validation} split of MSR-VTT.}
    Performance varies smoothly with inverse temperature,
    peaking at a value of 20.
    }
    \mbox{}\vspace{-0.7cm} \\
    \label{fig:num-queries-and-hyperparam}
\end{figure}

\noindent \textbf{What is the influence of querybank size on performance?}
To address this, we sample querybanks across a range of 
different scales, and report mean and standard deviations across
metrics for three samplings of each scale using DIS normalisation.
The results are shown in Fig.~\ref{fig:num-queries-and-hyperparam} (left),
where we observe that performance increases with querybank size,
but strong results can be obtained with a querybank of just
a few thousand random training samples.

\noindent \textbf{What is the influence of the similarity normalisation strategy on \methodName?}
To address this question,
we first sample querybanks of 5,000 samples from the MSR-VTT
training split and compare the normalisation strategies
described in Sec.~\ref{sec:method:framework-components}.
Results are reported in
the upper block (\textit{``In Domain''})
of Tab.~\ref{tab:querybank-domain-gap}
where we observe that CSLS~\cite{conneau2018word}, IS~\cite{smith2017offline}
and the proposed DIS strategy perform best,
and that all querybank normalisation methods 
\textit{substantially outperform the baseline without
normalisation}.
Next, to evaluate the robustness of the normalisation
strategies to different querybank sampling distributions,
we sample additional querybanks of 5,000 samples from
the training splits of two different video
retrieval datasets:
MSVD~\cite{Chen2011CollectingHP}
(whose query domain closely matches MSR-VTT),
and
LSMDC~\cite{Rohrbach2015ADF}
(a collection of movies with audio descriptions,
whose query domain is further away from MSR-VTT),
and evaluate retrieval performance on
MSR-VTT test.
We report results in the middle blocks of 
Tab.~\ref{tab:querybank-domain-gap}
(\textit{``Close Domain''} and \textit{``Far Domain''})
where we observe that sampling the querybank
from a closely overlapping domain (MSVD) works well
for all methods (with DIS performing best),
but that sampling from a different domain (LSMDC)
\textit{degrades performance below the
baseline without normalisation for all methods
except GC~\cite{dinu2014improving} and DIS}.

To understand why the LSMDC querybank could be actively
harmful for methods other than GC~\cite{dinu2014improving}
and DIS,
we studied the samples closely
and observed that LSMDC queries retrieve only a small
subset of videos from the video gallery
(and thus were ineffectual at their primary purpose of probing
for hubs).
To validate that this retrieval distribution was indeed the cause
of the issue, we constructed an ``adversarial'' querybank from
MSR-VTT by selecting the 5,000 training queries
that achieved the smallest coverage (i.e. retrieved the lowest
number of distinct videos) over the MSR-VTT test set. 
We report numbers in the \textit{Adversarial} block of Tab.~\ref{tab:querybank-domain-gap}. 
We observe that despite sampling from the same dataset,
all normalisation methods other than DIS are significantly harmed.
In the lower block, \textit{Overall}, we present the overall performance computed as geometric mean for all methods. Since DIS performs the best overall (presented in bold in Tab.~\ref{tab:querybank-domain-gap}), we
use it as our normalisation strategy for \methodName
for all remaining experiments.

\input{tables/querybank-domain-gap}

\begin{figure}
    \centering
    \includegraphics[trim={0cm 0.5cm 0cm 0.2cm},clip,width=\linewidth]{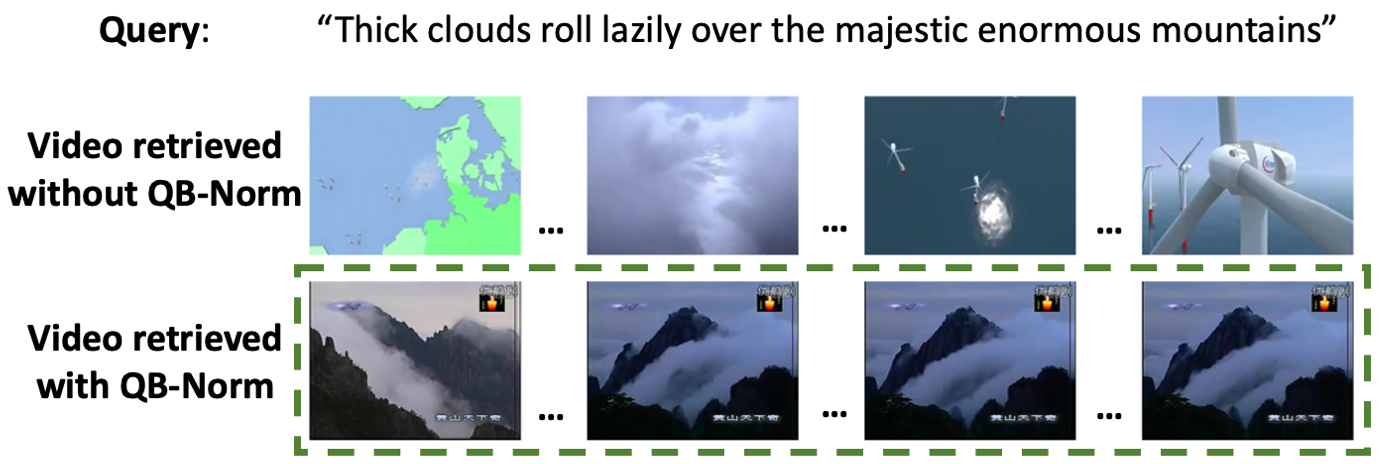}
    \vspace{-0.6cm}
    \caption{\textbf{Qualitative results.}
    We illustrate a sample query for which \methodName leads to the retrieval of the
    correct target video (whose frames are highlighted with a dashed green line).
    For further examples and more detailed analysis, see supplementary. 
    }
    \mbox{}\vspace{-0.7cm} \\
    \label{fig:qualitative}
\end{figure}

\noindent \textbf{Hyperparameter sensitivity.}
The IS~\cite{conneau2018word} and DIS normalisation strategies require
the user to select an additional hyperparameter (the inverse temperature)
that is absent from other methods.
We evaluate the sensitivity of \textit{DIS} to this hyperparameter in
Fig.~\ref{fig:num-queries-and-hyperparam} (right),
where we find that a value of 20 works best.
In practice, we found that this value worked well consistently
across datasets, and therefore we use it for all remaining 
experiments (with the exception of CLIP2Video \cite{fang2021clip2video} where we used
$1.99^{-1}$,
since the similarities are already scaled by the method).
DIS normalisation introduces an additional hyperparameter
(the $k$ maximum selection value described in Sec.~\ref{sec:method:framework-components}).
We observed that choosing $k=1$
offers a good trade-off between good performance and
robustness, so we simply use this value for all experiments.

\noindent \textbf{Does \methodName mitigate hubness?}
The core motivation for \methodName is that existing 
cross modal retrieval methods are heavily affected by hubness
(Fig.~\ref{fig:motivation}).
To investigate whether this has been addressed by \methodName,
we report the
\textit{skewness of the k-occurrences
distribution}%
\footnote{A detailed description of this
calculation is given in the supplementary.}
(which indicates the hubness of an embedding
space~\cite{radovanovic2010hubs})
for four datasets in Tab.~\ref{tab:hubness-mitigation}
using a querybank consisting from all the samples from the training set.
We observe that in each case,
skewness (and hence hubness) is \textit{significantly reduced}.

\input{tables/hubness_mitigation}

\subsection{Comparison with other methods}
\label{sec:experiments:benchmarks}

\input{tables/sota}
\input{tables/image_retrieval}
\input{tables/image_image}
\input{tables/audio_retrieval}

In this section, we conduct an extensive study to evaluate the 
effectiveness and generality of \methodName on several well established benchmarks.

The influence of applying \methodName to cross modal embeddings
for \textbf{text-video retrieval} are reported in
Tab.
~\ref{tab:msrvtt-1ka-sota},
~\ref{tab:msvd-final-sota},
~\ref{tab:didemo-final-sota},
~\ref{tab:lsmdc-final-sota},
~\ref{tab:vatex-final-sota},
~\ref{tab:queryd-final-sota}.
We provide further text-video retrieval results in the supplementary.
In Tab.~\ref{tab:mscoco-final-sota}
we report results for the \textbf{text-image retrieval} task, while in 
Tab.~\ref{tab:cub-final-sota},
~\ref{tab:online-final-sota}, we report results for the \textbf{image-image retrieval} task.
Finally, in Tab.~\ref{tab:audiocaps-final-sota},
we report results for \textbf{text-audio retrieval}. In Fig.~\ref{fig:qualitative} we also show a qualitative example. For the base models that provide weights for different seeds we report mean and standard deviation of \methodName applied on each seed.
In each case, \methodName brings a significant improvement over all tested methods, benchmarks and tasks. We show in bold the best performing method.

%% file: tables/querybank-train-vs-test.tex
\begin{table}
\begin{center}
\resizebox{\linewidth}{!}{
\begin{tabular}{c|c|c|c|c|c}%
\hline%
\hline%
Querybank Source&Size&$R@1\uparrow$&$R@5\uparrow$&$R@10\uparrow$&$MdR\downarrow$\\%
\hline%
\textbf{No querybank} & - &$14.9_{\pm0.1}$&$38.3_{\pm0.1}$&$51.5_{\pm0.1}$&$10.0_{\pm0.0}$\\%
\textbf{Training set} & 60k &$17.3_{\pm0.0}$&$42.1_{\pm0.1}$&$54.9_{\pm0.0}$&$8.0_{\pm0.0}$\\%
\textbf{Val set} & 10k & $16.6_{\pm0.1}$&$40.8_{\pm0.1}$&$53.7_{\pm0.1}$&$9.0_{\pm0.0}$\\%
\textbf{Test set} & 60k &$17.5_{\pm0.0}$&$42.4_{\pm0.1}$&$55.1_{\pm0.0}$&$8.0_{\pm0.0}$\\%
\hline%
\end{tabular}}
\end{center}
\vspace{\spacebefore{}}
\caption{\textbf{Effective querybanks can be constructed
from the training set.}
Performance is reported on MSR-VTT
\texttt{full} split~\cite{xu2016msr}.
We observe that a querybank of 60K samples from the training
set performs comparably to a test set querybank.
\label{tab:querybank-train-vs-test}}
\vspace{\spaceafter{}}
\end{table}

%% file: tables/querybank-domain-gap.tex
\setlength{\tabcolsep}{1pt}
\begin{table}
\begin{center}
\resizebox{\linewidth}{!}{
\begin{tabular}{c|c|c|c|c|c}%
\hline%
\hline%
\makecell{QB Source\\Data} & Normalisation & $R@1\uparrow$ & $R@5\uparrow$  & $R@10\uparrow$ &  $MdR\downarrow$\\%
\hline%
\textbf{No QB} & - &$14.9_{\pm0.1}$&$38.3_{\pm0.1}$&$51.5_{\pm0.1}$&$10.0_{\pm0.0}$\\%
\hline
\hline
\textit{In Domain} \\
\hline
\textbf{MSR-VTT}  & \methodName(GC~\cite{dinu2014improving}) &$15.8_{\pm0.0}$&$39.1_{\pm0.0}$&$51.8_{\pm0.0}$&$10.0_{\pm0.0}$\\%
\textbf{MSR-VTT} & \methodName(CSLS~\cite{conneau2018word}) &$16.8_{\pm0.1}$&$41.5_{\pm0.1}$&$54.4_{\pm0.1}$&$8.0_{\pm0.0}$\\%
\textbf{MSR-VTT}& \methodName(IS~\cite{smith2017offline}) & $17.1_{\pm0.1}$&$41.9_{\pm0.2}$&$54.7_{\pm0.1}$&$8.0_{\pm0.0}$\\%
\textbf{MSR-VTT}& \methodName(DIS) & $17.0_{\pm0.1}$&$41.3_{\pm0.1}$&$54.1_{\pm0.1}$&$8.6_{\pm0.5}$\\%
\hline
\hline
\textit{Close Domain} \\
\hline
\textbf{MSVD} & \methodName(GC~\cite{dinu2014improving}) &$15.2_{\pm0.1}$&$38.8_{\pm0.0}$&$51.7_{\pm0.0}$&$10.0_{\pm0.0}$\\%
\textbf{MSVD} & \methodName(CSLS~\cite{conneau2018word}) &$16.5_{\pm0.0}$&$41.2_{\pm0.0}$&$54.1_{\pm0.1}$&$9.0_{\pm0.0}$\\%
\textbf{MSVD} & \methodName(IS~\cite{smith2017offline}) & $16.4_{\pm0.2}$&$40.9_{\pm0.2}$&$53.9_{\pm0.1}$&$9.0_{\pm0.0}$\\%
\textbf{MSVD} & \methodName(DIS) &$16.7_{\pm0.1}$&$41.1_{\pm0.1}$&$54.0_{\pm0.0}$&$9.0_{\pm0.0}$\\%
\hline
\hline
\textit{Far Domain} \\
\hline
\textbf{LSMDC} & \methodName(GC~\cite{dinu2014improving}) &$14.8_{\pm0.1}$&$38.2_{\pm0.0}$&$51.4_{\pm0.0}$&$10.0_{\pm0.0}$\\%
\textbf{LSMDC} & \methodName(CSLS~\cite{conneau2018word}) &$13.4_{\pm0.0}$&$35.9_{\pm0.0}$&$48.5_{\pm0.0}$&$11.0_{\pm0.0}$\\%
\textbf{LSMDC} & \methodName(IS~\cite{smith2017offline}) &$11.6_{\pm0.0}$&$32.5_{\pm0.0}$&$44.6_{\pm0.0}$&$14.0_{\pm0.0}$\\%
\textbf{LSMDC} & \methodName(DIS) &$14.9_{\pm0.1}$&$38.3_{\pm0.1}$&$51.2_{\pm0.1}$&$10.0_{\pm0.0}$\\%
\hline
\hline
\textit{Adversarial} \\
\hline
\textbf{MSR-VTT} & \methodName(GC~\cite{dinu2014improving}) &$14.5_{\pm0.0}$&$38.1_{\pm0.0}$&$51.4_{\pm0.0}$&$10.0_{\pm0.0}$\\%
\textbf{MSR-VTT} & \methodName(CSLS~\cite{conneau2018word}) &$14.4_{\pm0.1}$&$37.5_{\pm0.1}$&$50.4_{\pm0.1}$&$10.0_{\pm0.0}$\\%
\textbf{MSR-VTT}& \methodName(IS~\cite{smith2017offline}) &$12.3_{\pm0.1}$&$32.9_{\pm0.1}$&$45.0_{\pm0.0}$&$14.0_{\pm0.0}$\\%
\textbf{MSR-VTT}& \methodName(DIS) &$14.9_{\pm0.1}$&$38.3_{\pm0.1}$&$51.5_{\pm0.1}$&$10.0_{\pm0.0}$\\%
\hline%
 \hline
 \textit{Overall} & & \makecell{$GM$ \\ $(R@1)$} & \makecell{$GM$ \\$(R@5)$} & \makecell{$GM$\\$(R@10)$} & \makecell{$GM$\\$(MdR)$} \\
 \hline
 \textbf{Summary}
 & \methodName(GC~\cite{dinu2014improving}) 
 & $15.1_{\pm0.6}$ & $38.5_{\pm0.5}$ & $51.6_{\pm0.2}$ & $10.0_{\pm0.0}$ \\%
 \textbf{Summary}
 & \methodName(CSLS~\cite{conneau2018word}) 
 & $15.2_{\pm1.6}$ & $39.0_{\pm2.8}$ & $51.8_{\pm2.9}$ & $\mathbf{9.4}_{\pm1.3}$ \\%
 \textbf{Summary}
 & \methodName(IS~\cite{smith2017offline}) 
 & $14.1_{\pm2.8}$ & $36.8_{\pm5.0}$ & $49.3_{\pm5.5}$ & $10.9_{\pm3.2}$ \\%
 \textbf{Summary}
 & \methodName(DIS)
 & $\mathbf{15.8}_{\pm1.1}$ & $\mathbf{39.7}_{\pm1.7}$ & $\mathbf{52.7}_{\pm1.6}$ & $\mathbf{9.4}_{\pm0.7}$ \\%
 \hline
\end{tabular}}
\end{center}
\vspace{\spacebefore{}}
\caption{\textbf{
The influence of normalisation strategies
across
querybank source distributions.} 
Performance is reported on MSR-VTT \texttt{full} split~\cite{xu2016msr},
while querybanks of 5,000 samples
are sampled from the training sets of different datasets.
In the last block, we presented the overall performance reported as geometric mean (GM) for each method.
We observe that DIS provides the best overall trade-off:
it matches the high performance of IS and CSLS
with \textit{in domain} and \textit{close domain}
querybanks, and is \textit{more robust} on \textit{far domain}
and \textit{adversarial} querybanks.
\label{tab:querybank-domain-gap}}
\vspace{\spaceafter{}}
\end{table}
\setlength{\tabcolsep}{6pt}

%% file: tables/hubness_mitigation.tex
\begin{table}
\begin{center}
\resizebox{0.9\linewidth}{!}{
\begin{tabular}{c|c|c|c|c|c|c|c}%
\hline%
\hline%
\multicolumn{2}{c}{MSR-VTT} & \multicolumn{2}{c}{DiDeMo} & \multicolumn{2}{c}{LSMDC} & \multicolumn{2}{c}{MSCoCo} \\%
Before & After & Before & After & Before & After & Before & After \\%
\hline%
0.939 & 0.509 & 1.21 & 0.39 & 0.715 & 0.321 & 0.56 & 0.16 \\
\hline%
\end{tabular}}
\end{center}
\vspace{\spacebefore{}}
\caption{\textbf{Impact of \methodName{} on hubness on various datasets.} We observe that \methodName consistently reduces hubness (as measured by skewness in the k-occurences distribution). \label{tab:hubness-mitigation}}
\vspace{\spaceafter{}}
\end{table}

%% file: tables/sota.tex
\begin{table}
\begin{center}
\resizebox{\linewidth}{!}{
\begin{tabular}{c|c|c|c|c}%
\hline%
\hline%
Model&$R@1\uparrow$&$R@5\uparrow$&$R@10\uparrow$&$MdR\downarrow$\\%
\hline%
CE\cite{liu2019use}&$21.7_{\pm1.3}$&$51.8_{\pm0.5}$&$65.7_{\pm0.6}$&$5.0_{\pm0.0}$\\
MMT\cite{gabeur2020multi}&$24.6_{\pm0.4}$&$54.0_{\pm0.2}$&$67.1_{\pm0.5}$&$4.0_{\pm0.0}$\\%
SSB\cite{patrick2020support}&$27.4$&$56.3$&$67.7$&$3.0$\\%
Frozen\cite{bain2021frozen}&$31.0$&$59.5$&$70.5$&$3.0$\\%
CLIP4Clip~\cite{luo2021clip4clip}&$44.5$&$71.4$&$81.6$&$\mathbf{2.0}$\\
\hline
TT-CE+\cite{croitoru2021teachtext}&$29.6_{\pm0.3}$&$61.6_{\pm0.5}$&$74.2_{\pm0.3}$&$3.0_{\pm0.0}$\\%
\textbf{TT-CE+ (+\methodName)} &$33.3_{\pm0.7}$&$63.7_{\pm0.1}$&$76.3_{\pm0.4}$&$3.0_{\pm0.0}$\\%
\hline
CLIP2Video\cite{fang2021clip2video} & $45.6$ & $72.5$ & $81.7$ & $\mathbf{2.0}$ \\
\textbf{CLIP2Video (+\methodName)}&$\mathbf{47.2}$&$\mathbf{73.0}$&$\mathbf{83.0}$&$\mathbf{2.0}$\\%
\hline%
\end{tabular}}
\end{center}
\vspace{\spacebefore{}}
\caption{\textbf{MSR-VTT 1k-A split: Comparison to state of the art.} \label{tab:msrvtt-1ka-sota}}
\vspace{\spaceafter{}}
\vspace{-0.5cm}
\end{table}

\begin{table}
\begin{center}
\resizebox{\linewidth}{!}{
\begin{tabular}{c|c|c|c|c}%
\hline%
\hline%
Model&$R@1\uparrow$&$R@5\uparrow$&$R@10\uparrow$&$MdR\downarrow$\\%
\hline%
VSE++\cite{faghri2017vse++}&$15.4$&$39.6$&$53.0$&$9.0$\\%
MoEE\cite{miech2018learning}&$21.1_{\pm0.2}$&$52.0_{\pm0.7}$&$66.7_{\pm0.2}$&$5.0_{\pm0.0}$\\%
CE\cite{liu2019use}&$21.5_{\pm0.5}$&$52.3_{\pm0.8}$&$67.5_{\pm0.7}$&$5.0_{\pm0.0}$\\%
Frozen\cite{bain2021frozen}&$33.7$&$64.7$&$76.3$&$3.0$\\%
CLIP4Clip~\cite{luo2021clip4clip}&$46.2$&$76.1$&$84.6$&$\mathbf{2.0}$\\
\hline%
TT-CE+\cite{croitoru2021teachtext}&$25.4_{\pm0.3}$&$56.9_{\pm0.4}$&$71.3_{\pm0.2}$&$4.0_{\pm0.0}$\\%
\textbf{TT-CE+ (+\methodName)}&$26.6_{\pm0.9}$&$58.5_{\pm1.3}$&$71.8_{\pm1.1}$&$4.0_{\pm0.0}$\\%
\hline%
CLIP2Video\cite{fang2021clip2video}&$47.0$&$76.8$&$85.9$&$\mathbf{2.0}$\\
\textbf{CLIP2Video (+\methodName)}&$\mathbf{47.6}$&$\mathbf{77.6}$&$\mathbf{86.1}$&$\mathbf{2.0}$\\%
\hline
\end{tabular}}
\end{center}
\vspace{\spacebefore{}}
\caption{\textbf{MSVD: Comparison to state of the art methods}. \label{tab:msvd-final-sota}}
\vspace{\spaceafter{}}
\end{table}

\begin{table}
\begin{center}
\resizebox{\linewidth}{!}{
\begin{tabular}{c|c|c|c|c}%
\hline%
\hline%
Model&$R@1\uparrow$&$R@5\uparrow$&$R@10\uparrow$&$MdR\downarrow$\\%
\hline%
MoEE\cite{miech2018learning}&$16.1_{\pm1.0}$&$41.2_{\pm1.6}$&$55.2_{\pm1.6}$&$8.3_{\pm0.5}$\\%
CE\cite{liu2019use}&$17.1_{\pm0.9}$&$41.9_{\pm0.2}$&$56.0_{\pm0.5}$&$8.0_{\pm0.0}$\\%
TT-CE&$21.0_{\pm0.6}$&$47.5_{\pm0.9}$&$61.9_{\pm0.5}$&$6.0_{\pm0.0}$\\%
Frozen\cite{bain2021frozen}&$31.0$&$59.8$&$72.4$&$3.0$\\%
CLIP4Clip~\cite{luo2021clip4clip}&$\mathbf{43.4}$&$\mathbf{70.2}$&$\mathbf{80.6}$&$\mathbf{2.0}$\\
\hline%
CE+\cite{croitoru2021teachtext}&$18.2_{\pm0.2}$&$43.9_{\pm0.9}$&$57.1_{\pm0.8}$&$7.9_{\pm0.1}$\\%
\textbf{CE+ (+\methodName)}&$20.7_{\pm0.6}$&$46.6_{\pm0.2}$&$59.8_{\pm0.2}$&$6.3_{\pm0.5}$\\%
\hline
TT-CE+\cite{croitoru2021teachtext}&$21.6_{\pm0.7}$&$48.6_{\pm0.4}$&$62.9_{\pm0.6}$&$6.0_{\pm0.0}$\\%
\textbf{TT-CE+ (+\methodName)}&$24.2_{\pm0.7}$&$50.8_{\pm0.7}$&$64.4_{\pm0.1}$&$5.3_{\pm0.5}$\\%
\hline
\hline
\end{tabular}}
\end{center}
\vspace{\spacebefore{}}
\caption{\textbf{DiDeMo: Comparison to state of the art methods}.%
\label{tab:didemo-final-sota}}
\vspace{\spaceafter{}}
\end{table}

\begin{table}
\begin{center}
\resizebox{\linewidth}{!}{
\begin{tabular}{c|c|c|c|c}%
\hline%
\hline%
Model&$R@1\uparrow$&$R@5\uparrow$&$R@10\uparrow$&$MdR\downarrow$\\%
\hline%
MoEE\cite{miech2018learning}&$12.1_{\pm0.7}$&$29.4_{\pm0.8}$&$37.7_{\pm0.2}$&$23.2_{\pm0.8}$\\%
CE\cite{liu2019use}&$12.4_{\pm0.7}$&$28.5_{\pm0.8}$&$37.9_{\pm0.6}$&$21.7_{\pm0.6}$\\%
MMT\cite{gabeur2020multi}&$13.2_{\pm0.4}$&$29.2_{\pm0.8}$&$38.8_{\pm0.9}$&$21.0_{\pm1.4}$\\%
Frozen\cite{bain2021frozen}&$15.0$&$30.8$&$39.8$&$20.0$\\%
CLIP4Clip~\cite{luo2021clip4clip}&$\mathbf{21.6}$&$\mathbf{41.8}$&$\mathbf{49.8}$&$\mathbf{11.0}$\\
\hline%
CE+\cite{croitoru2021teachtext}&$14.9_{\pm0.6}$&$33.7_{\pm0.2}$&$44.1_{\pm0.6}$&$15.3_{\pm0.5}$\\%
\textbf{CE+ (\methodName)}&$16.4_{\pm0.8}$&$34.8_{\pm0.4}$&$44.9_{\pm0.9}$&$14.5_{\pm0.4}$\\%
\hline
TT-CE+\cite{croitoru2021teachtext}&$17.2_{\pm0.4}$&$36.5_{\pm0.6}$&$46.3_{\pm0.3}$&$13.7_{\pm0.5}$\\%
\textbf{TT-CE+ (\methodName)}&$17.8_{\pm0.4}$&$37.7_{\pm0.5}$&$47.6_{\pm0.6}$&$12.7_{\pm0.5}$\\%
\hline%
\end{tabular}}
\end{center}
\vspace{\spacebefore{}}
\caption{\textbf{LSMDC: Comparison to state of the art methods}.%
\label{tab:lsmdc-final-sota}}
\vspace{\spaceafter{}}
\end{table}

\begin{table}
\begin{center}
\resizebox{\linewidth}{!}{
\begin{tabular}{c|c|c|c|c}%
\hline%
\hline%
Model&$R@1\uparrow$&$R@5\uparrow$&$R@10\uparrow$&$MdR\downarrow$\\%
\hline%
HGR\cite{chen2020fine}&$35.1$&$73.5$&$83.5$&$2.0$\\%
SSB\cite{patrick2020support}&$44.6$&$81.8$&$89.5$&$\mathbf{1.0}$\\%
CE\cite{liu2019use}&$47.9_{\pm0.1}$&$84.2_{\pm0.1}$&$91.3_{\pm0.1}$&$2.0_{\pm0.0}$\\%
Fast and Slow~\cite{miech2021thinking}&$50.5$&$84.6$&$91.7$&-\\
\hline
TT-CE+\cite{croitoru2021teachtext}&$53.2_{\pm0.2}$&$87.4_{\pm0.1}$&$93.3_{\pm0.0}$&$\mathbf{1.0}_{\pm0.0}$\\%
\textbf{TT-CE+ (+\methodName)}&$54.8_{\pm0.1}$&$88.2_{\pm0.1}$&$\mathbf{93.8}_{\pm0.1}$&$\mathbf{1.0}_{\pm0.0}$\\%
\hline%
CLIP2Video\cite{fang2021clip2video}&$57.4$&$87.9$&$93.6$&$\mathbf{1.0}$\\%
\textbf{CLIP2Video (+\methodName)} &$\mathbf{58.8}$&$\mathbf{88.3}$&$\mathbf{93.8}$&$\mathbf{1.0}$\\%
\hline%
\end{tabular}}
\end{center}
\vspace{\spacebefore{}}
\caption{\textbf{VaTeX: Comparison to state of the art methods}. \label{tab:vatex-final-sota}}
\vspace{\spaceafter{}}
\end{table}

\begin{table}
\begin{center}
\resizebox{\linewidth}{!}{
\begin{tabular}{c|c|c|c|c}%
\hline%
\hline%
Model&$R@1\uparrow$&$R@5\uparrow$&$R@10\uparrow$&$MdR\downarrow$\\%
\hline%
MoEE\cite{miech2018learning}&$11.6_{\pm1.3}$&$30.2_{\pm3.0}$&$43.2_{\pm3.1}$&$14.2_{\pm1.6}$\\%
CE\cite{liu2019use}&$13.9_{\pm0.8}$&$37.6_{\pm1.2}$&$48.3_{\pm1.4}$&$11.3_{\pm0.6}$\\%
\hline%
CE+\cite{croitoru2021teachtext}&$13.2_{\pm2.0}$&$37.1_{\pm2.9}$&$50.5_{\pm1.9}$&$10.3_{\pm1.2}$\\%
\textbf{CE+ (+\methodName)}&$14.1_{\pm1.8}$&$\mathbf{38.6}_{\pm1.3}$&$51.1_{\pm1.6}$&$10.0_{\pm0.8}$\\%
\hline
TT-CE+\cite{croitoru2021teachtext}&$14.4_{\pm0.5}$&$37.7_{\pm1.7}$&$50.9_{\pm1.6}$&$\mathbf{9.8}_{\pm1.0}$\\%
\textbf{TT-CE+ (+\methodName)}&$\mathbf{15.1}_{\pm1.6}$&$38.3_{\pm2.4}$&$\mathbf{51.2}_{\pm2.8}$&$10.3_{\pm1.7}$\\%
\hline%
\end{tabular}}
\end{center}
\vspace{\spacebefore{}}
\caption{\textbf{QuerYD: Comparison to state of the art methods}. \label{tab:queryd-final-sota}}
\vspace{\spaceafter{}}
\end{table}

%% file: tables/image_retrieval.tex
\begin{table}
\begin{center}
\resizebox{\linewidth}{!}{
\begin{tabular}{c|c|c|c|c}%
\hline%
\hline%
Model&$R@1\uparrow$&$R@5\uparrow$&$R@10\uparrow$&$MdR\downarrow$\\%
\hline%
CLIP~\cite{radford2021learning} & $37.8$ & $62.4$ & $72.2$ & - \\
VSE++~\cite{faghri2017vse++}&$43.9$ &$59.4$&$72.4$&- \\
OSCAR~\cite{li2020oscar}&$54.0$&$80.8$&$88.5$&-\\ %
VinVL~\cite{zhang2021vinvl}&$58.8$&$83.5$&$90.3$&- \\
Fast and Slow~\cite{miech2021thinking}&$\mathbf{68.2}$&$\mathbf{89.7}$&$\mathbf{93.9}$&- \\
\hline
CLIP\cite{radford2021learning}$\ssymbol{3}$&$30.3$&$56.1$&$67.1$&$4.0$\\
\textbf{CLIP$\ssymbol{3}$ (+\methodName)}&$34.8$&$59.9$&$70.4$&$3.0$\\%
\hline%
MMT-OSCAR~\cite{geigle2021retrieve}&$52.2$&$80.2$&$88.0$&$\mathbf{1.0}$\\%
\textbf{MMT-Oscar (+\methodName)}&$53.9$&$80.5$&$88.1$&$\mathbf{1.0}$\\%
\hline
\end{tabular}}
\end{center}
\vspace{\spacebefore{}}
\caption{\textbf{Text-image retrieval - MSCoCo 5k split: Comparison to other methods.} $\ssymbol{3}$ represents the results obtained using the official CLIP\cite{radford2021learning} ViT-B/32 model. \label{tab:mscoco-final-sota}}
\vspace{\spaceafter{}}
\end{table}

%% file: tables/image_image.tex
\begin{table}
\begin{center}
\resizebox{\linewidth}{!}{
\begin{tabular}{c|c|c|c|c}%
\hline%
\hline%
Model&$R@1\uparrow$&$R@2\uparrow$&$R@4\uparrow$&$R@8\uparrow$\\%
\hline%
MS~\cite{wang2019multi} & $57.4$ & $69.8$ & $80.0$ & - \\
EPS~\cite{levi2021rethinking} & $64.4$ & $75.2$ & $\mathbf{84.3}$ & - \\
\hline
RDML\cite{roth2020revisiting}& $64.4$ & $75.3$ & $83.4$ & $90.0$ \\
\textbf{RDML\cite{roth2020revisiting} (+\methodName)}& $\mathbf{64.8}$ & $\mathbf{75.6}$ & $84.0$ & $\mathbf{90.4}$ \\
\hline%
\end{tabular}}
\end{center}
\vspace{\spacebefore{}}
\caption{\textbf{Image to Image retrieval - CUB 200: Comparison to other methods.} \label{tab:cub-final-sota}}
\vspace{\spaceafter{}}
\end{table}

\begin{table}
\begin{center}
\resizebox{\linewidth}{!}{
\begin{tabular}{c|c|c|c|c}%
\hline%
\hline%
Model&$R@1\uparrow$&$R@10\uparrow$&$R@100\uparrow$&$R@1000\uparrow$\\%
\hline%
XBM~\cite{wang2020cross} & $\mathbf{80.6}$ & $\mathbf{91.6}$ & $96.2$ & $98.7$ \\
Smooth-AP\cite{brown2020smooth} & $80.1$ & $91.5$ & $\mathbf{96.6}$ & $\mathbf{99.0}$ \\
\hline
RDML\cite{roth2020revisiting} & $77.8$ & $89.5$ & $95.4$ & $98.4$ \\
\textbf{RDML\cite{roth2020revisiting} (+\methodName)}& $78.1$ & $89.8$ & $95.6$ & $98.5$\\
\hline%
\end{tabular}}
\end{center}
\vspace{\spacebefore{}}
\caption{\textbf{Image to Image retrieval - Online Products: Comparison to other methods.} \label{tab:online-final-sota}}
\vspace{\spaceafter{}}
\end{table}

%% file: tables/audio_retrieval.tex
\begin{table}
\begin{center}
\resizebox{\linewidth}{!}{
\begin{tabular}{c|c|c|c|c}%
\hline%
\hline%
Model&$R@1\uparrow$&$R@5\uparrow$&$R@10\uparrow$&$MdR\downarrow$\\%
\hline%
AR~\cite{oncescu2021audio}-MoEE &$22.5_{\pm0.3}$&$54.4_{\pm0.6}$&$69.5_{\pm0.9}$&$5.0_{\pm0.0}$\\%
\hline
AR~\cite{oncescu2021audio}-CE &$23.1_{\pm0.6}$&$55.1_{\pm0.7}$&$70.7_{\pm0.6}$&$4.7_{\pm0.5}$\\%
\textbf{AR~\cite{oncescu2021audio}-CE (+\methodName)} &$\mathbf{23.9}_{\pm0.2}$&$\mathbf{57.1}_{\pm0.3}$&$\mathbf{71.6}_{\pm0.4}$&$\mathbf{4.0}_{\pm0.0}$\\%
\hline%
\end{tabular}}
\end{center}
\vspace{\spacebefore{}}
\caption{\textbf{Text-audio retrieval - AudioCaps: Comparison to other methods.} \label{tab:audiocaps-final-sota}}
\vspace{-0.6cm}
\end{table}

%% file: 05-conclusions.tex
\section{Limitations and societal impact}
\label{sec:limitations}
\noindent \textbf{Limitations}
All the normalisation techniques used with \methodName incur additional pre-computation costs.
The proposed normalisation technique, DIS, adds a further small additional computational cost over other normalisation approaches. %
For a full discussion on complexity, please refer to the supplementary.
We also show in Tab.~\ref{tab:querybank-domain-gap}
that adversarial querybank selection
and significant domain gaps can reduce the benefits of \methodNameLong.

\noindent \textbf{Societal impact} Cross modal retrieval is a powerful tool with both positive
applications and risks of harm.
Cross modal search enables efficient content discovery for
researchers, musicians, artists and consumers.
However, this capability also lends itself to tools of political
oppression: for example, it could enable efficient searching
of social media content to discover signs of political dissent.

\section{Conclusions}
\label{sec:conclusion}
In this work, we introduced the \methodNameLong framework
for hubness mitigation.
We also proposed the Dynamic Inverted Softmax for
robust similarity normalisation.
We demonstrated its broad applicability across a range
of tasks, models and benchmarks.

%% file: 06-ack.tex
\vspace{0.2cm}
\\
\noindent\textbf{Acknowledgements.}
This work was supported by Adobe, Google and Zhejiang Lab (NO. 2022NB0AB05), and a G-Research travel grant.
The authors thank Bruno Korbar for useful suggestions, Andrew Zisserman and Jenny Hu for their support, and Adam Berenzweig for kindly sharing a copy of his earlier work.
S.A.\ would like to acknowledge Z.\ Novak, N.\ Novak and S.\ Carlson
in supporting his contribution.

%% file: supp-mat.tex
\appendix
\begin{center}
\textbf{\Large Appendix}
\end{center}

In this appendix, we provide additional information and ablation studies
relating to \methodName.
We begin by providing more details about the datasets used
for each task (Sec.~\ref{sec:dataset-detils}).
We then provide ablation studies that investigate:
(1) The influence of the $k$ hyperparameter on the proposed
DIS normalisation scheme (Sec.~\ref{sec:topk-supp});
(2) Whether effective querybanks can also be constructed from the
training set using IS normalisation,
rather than DIS normalisation (Sec.~\ref{sec:effective-is});
(3) How embedding dimensionality influences the effectiveness
of \methodName (Sec.~\ref{sec:embd-effect}).
Next, we present comparisons on additional datasets for the text-video retrieval task (Sec.~\ref{additional}). 
In Sec.~\ref{sec:complexity} we discuss the
complexity of each normalisation technique.
In Sec.~\ref{sec:cent} we present a comparison
with CENT~\cite{suzuki2013centering} normalisation.
Then, we provide details on the \textit{skewness}
metric reported in the submission (Sec.~\ref{sec:skewness}),
offer a more complete set of metrics
across ablations (Sec.~\ref{sec:other-metrics})
and
give more details about the text and video experts used in
this work (Sec.~\ref{sec:experts}).
Finally, we report metrics indicating how \methodName performs on
video-text retrieval (Sec.~\ref{sec:t2v})
and provide some additional qualitative
results (Sec.~\ref{sec:qualitative}).

\section{Dataset details}
\label{sec:dataset-detils}
In this section, we describe the splits and datasets employed for all tasks considered in this work.

\subsection{Text-video retrieval}
For the task of text-video retrieval we test our approach on seven current benchmarks.

\indent \textbf{MSR-VTT}~\cite{xu2016msr} contains around
10k videos, each having 20 captions.
For the task of text-video retrieval,
we follow prior works~\cite{liu2019use, croitoru2021teachtext}
and we report results on the official split (\texttt{full}) which
contains 2,990 videos for testing and 497 for validation.
Since a number of recent works
\cite{liu2019use,gabeur2020multi,patrick2020support,croitoru2021teachtext} 
also report results on the 1k-A split,
we compare against these method on this split as well.
The 1k-A split contains 1,000 videos for testing
and around 9,000 for training.
We use the same videos and captions as defined in~\cite{liu2019use} which are used by other works~\cite{yu2018joint,gabeur2020multi,patrick2020support} for evaluation.
We report the results using models trained for 100 epochs.

\indent \textbf{MSVD}~\cite{chen2011collecting} has
1,970 videos and around 80k captions.
We report results on the standard split using in prior
works~\cite{venugopalan2015sequence,xu2015jointly,liu2019use,croitoru2021teachtext}
which consists of 1,200 videos for training, 100 for validation and 670 for testing. 

\indent \textbf{DiDeMo}~\cite{anne2017localizing} has 10,464 videos.
They are collected from a large-scale creative commons
collection~\cite{thomee2016yfcc100m} and are varied in content
(concerts, sports, pets etc.).
For each video, there are 3-5 pairs of descriptions.
For the task of text-video retrieval,
we use the paragraph video retrieval protocol as defined in
prior works~\cite{zhang2018cross,liu2019use,croitoru2021teachtext}.
This means that we the split consisting of 8,392 for training,
1,065 validation and 1,004 test videos.

\indent \textbf{LSMDC}~\cite{rohrbach2017movie}
contains 118,081 short video clips extracted from 202 movies.
Each clip has a textual description which consist in a caption
which is extracted either from the movie script or transcribed
from descriptive video services (DVS) for the visually impaired.
We use the official splits as defined in the Large Scale Movie
Description Challenge (LSMDC). The testing split contains 1,000 videos.

\indent \textbf{VaTeX}~\cite{wang2019vatex} contains 3,4911 videos and
has multilingual captions in Chinese and English.
Each video has 10 captions for each language.
As for the other datasets, we follow the same protocol as defined
in prior works~\cite{chen2020fine,patrick2020support,croitoru2021teachtext}
and use 1,500 videos for testing, while there are 1,500 videos for validation.
Please note that in this work, we use only the English annotations.

\indent \textbf{QuerYD}~\cite{oncescu20queryd} has 1,815 videos for training,
388 for validation and 390 for testing.
The videos are extracted from YouTube and are varied in content.
The dataset has 31,441 textual descriptions.
13,019 of these are precisely localized in the video with start time and end
time annotations while the other 18,422 are coarsely localized.
In this work, we do not use the localization annotations and report results
on the official splits following prior work on
text-video retrieval~\cite{croitoru2021teachtext}.

\indent \textbf{ActivityNet}~\cite{caba2015activitynet} contains
20k videos and has around 100K descriptive sentences.
The videos are extracted from YouTube.
We use a paragraph video retrieval as defined in
prior works~\cite{zhang2018cross, liu2019use, croitoru2021teachtext}.
We report results on the \texttt{val1} split.
The training split consists of 10,009 videos, while there are 4,917
videos for testing.

\subsection{Text-image retrieval}

For text image retrieval, we report results on the \textbf{MSCoCo}~\cite{Chen2015MicrosoftCC} dataset. It consists of 123k images with 5 captions for each sentence. We report results for the 5k test split.

\subsection{Text-audio retrieval}

For text audio retrieval,
we report results on the \textbf{AudioCaps}~\cite{Kim2019AudioCapsGC} dataset
which comprises sounds with event descriptions.
We use the same setup as prior work~\cite{oncescu2021audio}
where 49,291 samples are used for training, 428 for validation and 816 for testing.

\subsection{Image-to-image retrieval}

\indent \textbf{CUB-200-2011}~\cite{Wah2011TheCB} contains
11,788 images with 200 classes.
The training split consist of the first 100 classes
(5,863 images) while the testing split contains the remaining
classes (5,924 images).
We use the same setup as used in prior work~\cite{roth2020revisiting}.

\indent \textbf{Stanford Online Products}~\cite{Song2016DeepML}
contains 120,053 images with products from 22,634 classes.
We use the provided train and test splits containing 59,551
and 60,502 images respectively,
as used in prior works~\cite{Song2016DeepML, roth2020revisiting}.

\section{The influence of the Top-k hyperparameter on DIS normalisation}
\label{sec:topk-supp}
\input{tables/topk-activations}
In Tab.~\ref{tab:topk-activations} we show the influence of $k$ in the Top-k selection employed when constructing the gallery activation set (introduced in Sec.~3.4 of the main paper).
We observe that choosing $k=1$ offers a 
good trade-off between good performance when
constructing \textit{In Domain} querybanks
and robustness when constructing
\textit{Far Domain} querybanks.
We therefore use $k=1$ for all reported experiments.

\section{Can effective querybanks can be constructed from the training set with IS normalisation?}
\label{sec:effective-is}
\input{tables/querybank-train-vs-test-is}
In the main submission, we showed that effective querybanks can be constructed
from the training set when employing DIS normalisation.
Here, we show that this property also applies to IS normalisation,
supporting our hypothesis that \methodNameLong has the general property of not requiring
concurrent access to multiple test queries for appropriate normalisation
strategies.
In Tab.~\ref{tab:querybank-train-vs-test-is} we report the results of
selecting queries from training, validation or testing split to form
the querybank when employing IS normalisation.
Similarly to DIS, we observe that training set querybanks perform
comparably to test set querybanks for IS normalisation.

\section{The influence of embedding dimensionality on the effectiveness of \methodName}
\label{sec:embd-effect}

\begin{figure}
    \centering
    \includegraphics[trim={0cm 0.5cm 0cm 0.2cm},clip,width=\linewidth]{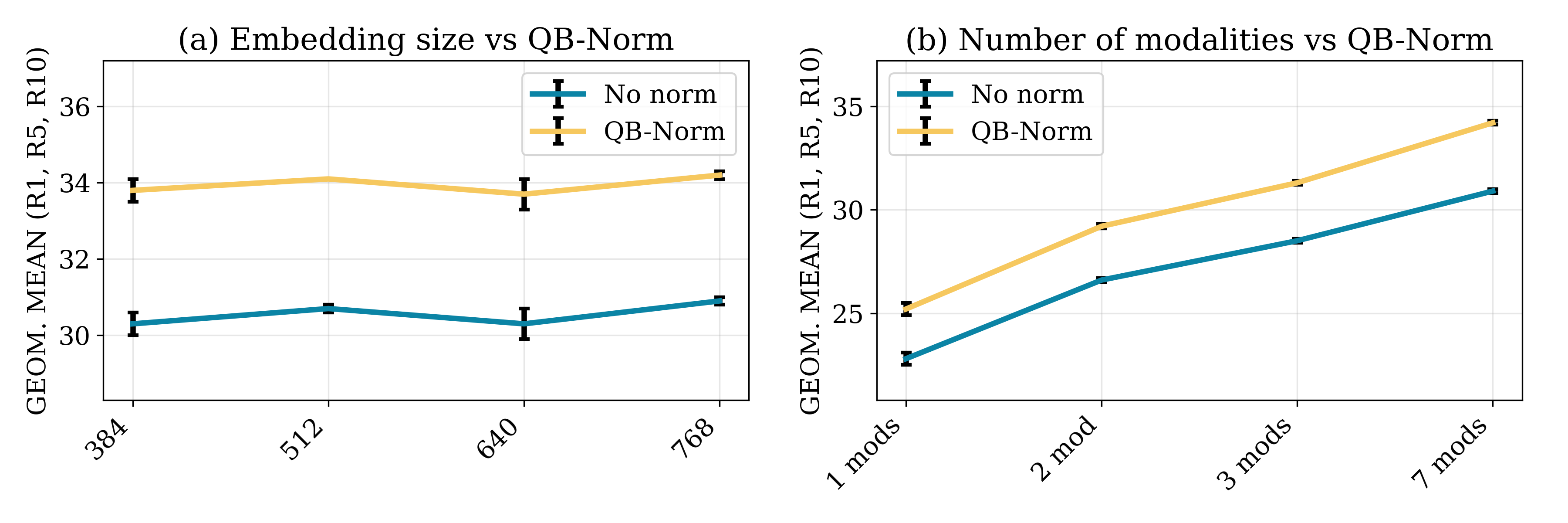}
    \caption{
    \textit{(Left)}:
    \textbf{The influence of embedding dimension on \methodName effectiveness}.
    We observe that \methodName brings a large increase in performance in all cases
    \textit{(Right)}:
    \textbf{The influence of number of used video embeddings on \methodName effectiveness.}
    We observe that our method is more effective with an increased number of modalities.
    }
    \mbox{}\vspace{-0.7cm} \\
    \label{fig:embedding-size}
\end{figure}

Radovanovic et al.~\cite{radovanovic2010hubs} posit that hubness
is a phenomenon that is:
(i) inherent to high dimensional spaces;
(ii) heavily influenced by the
\textit{intrinsic dimensionality} of the data.
To investigate these perspectives, we study the improvement
yielded by \methodName over embeddings of different dimensionality,
reporting results in Fig.~\ref{fig:embedding-size} (left).
We observe that \methodName brings around the same gain
when changing the embedding size.
We can interpret this finding within the framework
of~\cite{radovanovic2010hubs} as making the statement
that changing the shared embedding dimensionality
\textit{does not} influence intrinsic dimensionality.
To provide further analysis, 
we make a crude approximation to increasing/decreasing
intrinsic dimensionality by increasing/decreasing the
number of modalities employed in the video embedding.
Intuitively, since audio provides a different ``view'' of a sample
to visual data,
we expect a joint embedding with access to more modalities
to exhibit higher intrinsic dimensionality than one with only visual
cues.
We plot the effect of these changes
in Fig.~\ref{fig:embedding-size} (right).
We observe a slight increase in performance gain when
applying \methodName with an increased number of modalities,
which accords with the Radovanovic~\cite{radovanovic2010hubs} hypothesis.

\begin{figure*}
    \centering
    \includegraphics[trim={0cm 0.5cm 0cm 0.2cm},clip,width=\linewidth]{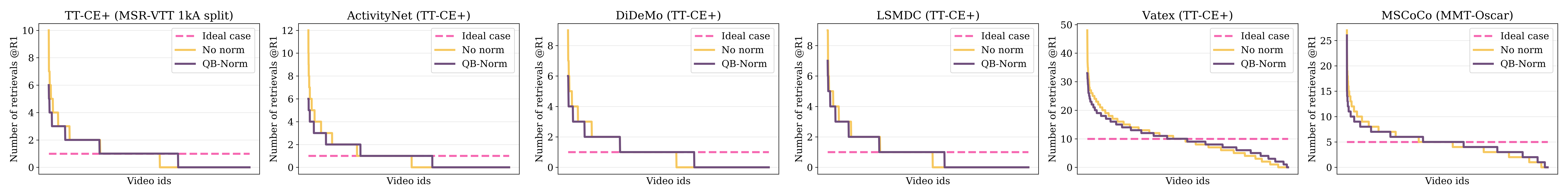}
    \vspace{-0.3cm}
    \caption{\textbf{Distribution of number of times each video is retrieved before and after applying \methodName.} We observe that \methodName reduces the maximum number of retrievals for any individual video. Furthermore, we note that with \methodName, previously unretrieved videos become possible to retrieve.}
    \mbox{}\vspace{-0.7cm} \\
    \label{fig:hubness-before-after}
\end{figure*}

\section{Additional text-video retrieval results}
\label{additional}
In Tab.~\ref{tab:msrvtt-final-sota},~\ref{tab:activity-net-final-sota} we report additional comparisons with state of the art on the MSR-VTT \texttt{full} split as well as ActivityNet~\cite{caba2015activitynet}.
In both cases, we observe that \methodName yields improvements.
We also explore the use of \methodName with CLIP4Clip~\cite{luo2021clip4clip}---for this, we train models using the code made available by the authors.
For CLIP4Clip experiments, we use a $\beta$ value of $0.45$ with the exception of LSMDC where $\beta$ is $1.26^{-1}$.
\input{tables/supp-sota}

\section{The computational complexity of normalisation strategies}
\label{sec:complexity}
As discussed in the main paper in Sec.3.4,
we use various normalization techniques in conjunction with \methodName.
In this section, we describe the computational cost of each technique
in the context of its influence on inference time.
For clarity of exposition,
we consider exact similarity searches,
but note that in practice approximate nearest neighbour implementations
are employed for large-scale deployments~\cite{johnson2019billion}.
All strategies incur an initial cost that corresponds to
pre-computing the similarity between a test query and
all the videos from the gallery,
$\mathcal{O}(N)$,
where $N$ represents the number of videos in the gallery.
We further assume that we have pre-computed and
stored all similarities between each query in the
querybank and videos from the gallery.
This assumption incurs both computational
and storage costs of
$\mathcal{O}(NM)$, where $M$ represents the number of
queries in the querybank.

\indent \textit{Globally-Corrected (GC) retrieval}~\cite{dinu2014improving}
involves determining the rank of the test query with respect
to the querybank for each gallery item.
Since we assume that we have pre-computed similarities between
the querybank and the gallery, we also pre-compute an initial
ranking over querybank elements for each gallery item.
For each test query, we establish its rank amongst the querybank
for every target item by performing a binary search over the
sorted list of pre-computed similarities.
This incurs an inference time cost of $\mathcal{O}(N \log M)$.

\indent \textit{Cross-Domain Similarity Local
Scaling (CSLS)}~\cite{conneau2018word}
consists of finding the most similar queries
from the querybank for each gallery video
and finding the $K$ gallery videos
(here $K$ is a hyperparameter of CSLS)
that are most similar to the test query.
For the former, we can pre-compute, for each video in the gallery,
the $K$ most similar queries from the querybank and store the
average similarity into a vector of size $N$.
For the latter,
we must compute (during inference)
the average similarity of the $K$ most similar
items among the gallery to our test query.
Using \textit{quickselect}, this can be done in $\mathcal{O}(N)$ time on
average (note that we do not require the top $K$ element similarities
to be sorted, since they will be averaged).

\indent \textit{Inverted Softmax (IS)}~\cite{smith2017offline}
involves normalizing the final similarity by the sum of the
similarities given the querybank.
However, the softmax denominator can be pre-computed by summing
the querybank similarities for each gallery item and storing the results
into a vector of size $N$.
During inference the similarities are divided by this pre-computed sum,
which adds only constant-time overhead.
Pre-computing the sum in this manner
also reduces the storage cost associated with the querybank
from $\mathcal{O}(NM)$ to $\mathcal{O}(N)$
(since we can discard the memory allocated to store the similarities
between each query in the querybank and each video in the gallery).

\indent \textit{Dynamic Inverted Softmax} (DIS).
Since DIS involves applying IS dynamically,
the computation of the normalization for each test query
is done in constant time as described above for IS.
The additional gallery activation set employed by DIS
can be pre-computed and stored for an additional
$\mathcal{O}(N)$ storage cost.
There is an additional cost during inference:
the top-$1$ search to determine the video originally retrieved
by the test query (which determines whether normalisation is performed).
This can be done in linear time ($\mathcal{O}(N)$).

\section{Comparison to CENT}
\label{sec:cent}
\input{tables/cent}

In Tab.~\ref{tab:msrvtt-cent}
we show how CENT~\cite{suzuki2013centering} normalisation performs
in comparison to an unnormalised baseline and \methodNameLong with DIS.
Since we found CENT to consistently harm performance for cross-modal
retrieval, we did not include it in all experiments in the main paper.

\section{Hubness and Skewness}
\label{sec:skewness}
We use the skewness metric as defined in~\cite{radovanovic2010hubs} to measure hubness:
\begin{equation}
    S_{N_k}=\frac{E(N_k - \mu_{N_k})^3}{\sigma_{N_k}^3}
\end{equation}

where $\mu_{N_k}$ and $\sigma_{N_k}$ are the mean and standard deviation of $N_k$. $N_k$ represents the k-occurrence distribution and is defined as follows $N_k(\mathbf{x}) = \sum_{i=1}^n p_{i,k}(\mathbf{x})$ where 
\begin{equation}
    p_{i,k}(\mathbf{x})=
    \begin{cases}
    1, \text{if $\mathbf{x}$ is among the $k$ nearest neighbours of $q_i$} \\
    0, \text{otherwise.}
    \end{cases}
\end{equation}

Here $\mathbf{x}$ represents a video embedding and $q_i \in Q$ a set of queries.
To compute these statistics,
In practice, we use the we use $k=10$, following~\cite{Feldbauer2018FastAH}
for the k-occurences distribution, employing the implementation of~\cite{Feldbauer2020}.

As shown in Tab. 3 in the main paper, skewness and hence hubness is reduced after applying \methodName. The same can be seen in Fig.~\ref{fig:hubness-before-after} which depicts the distribution of number of times each video is retrieved before and after using \methodName.
We observe that the maximum number of times a video
is retrieved is reduced, indicating a hubness reduction.

\section{Additional ablations on other metrics}
\label{sec:other-metrics}
\begin{figure}
    \centering
    \begin{subfigure}[b]{0.5\textwidth}
        \includegraphics[trim={0cm 0.5cm 0cm 0.2cm},clip,width=0.95\linewidth]{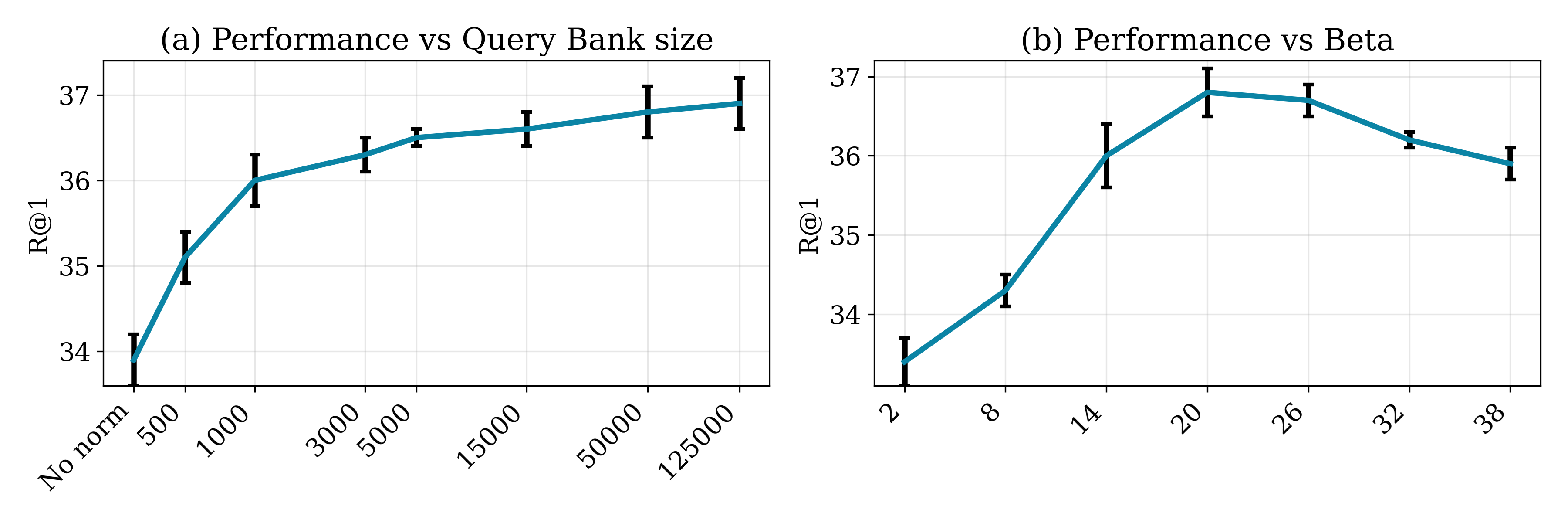}
    \end{subfigure}
    \begin{subfigure}[b]{0.5\textwidth}
        \includegraphics[trim={0cm 0.5cm 0cm 0.2cm},clip,width=0.95\linewidth]{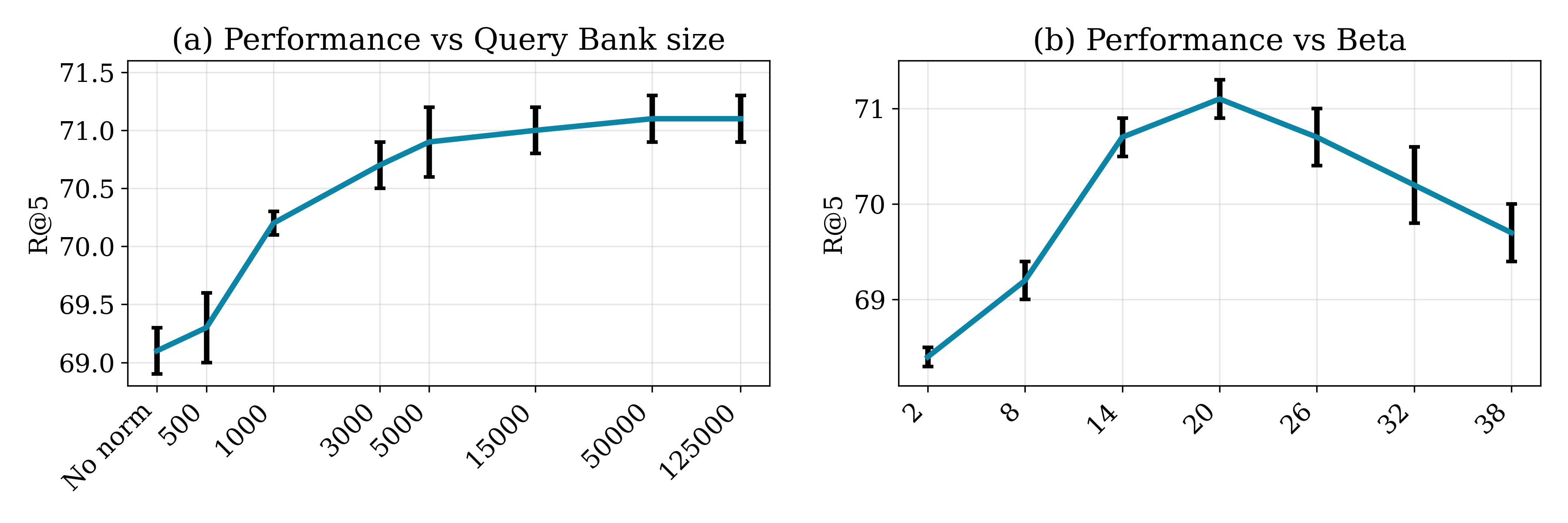}
    \end{subfigure}
    \begin{subfigure}[b]{0.5\textwidth}
        \includegraphics[trim={0cm 0.5cm 0cm 0.2cm},clip,width=0.95\linewidth]{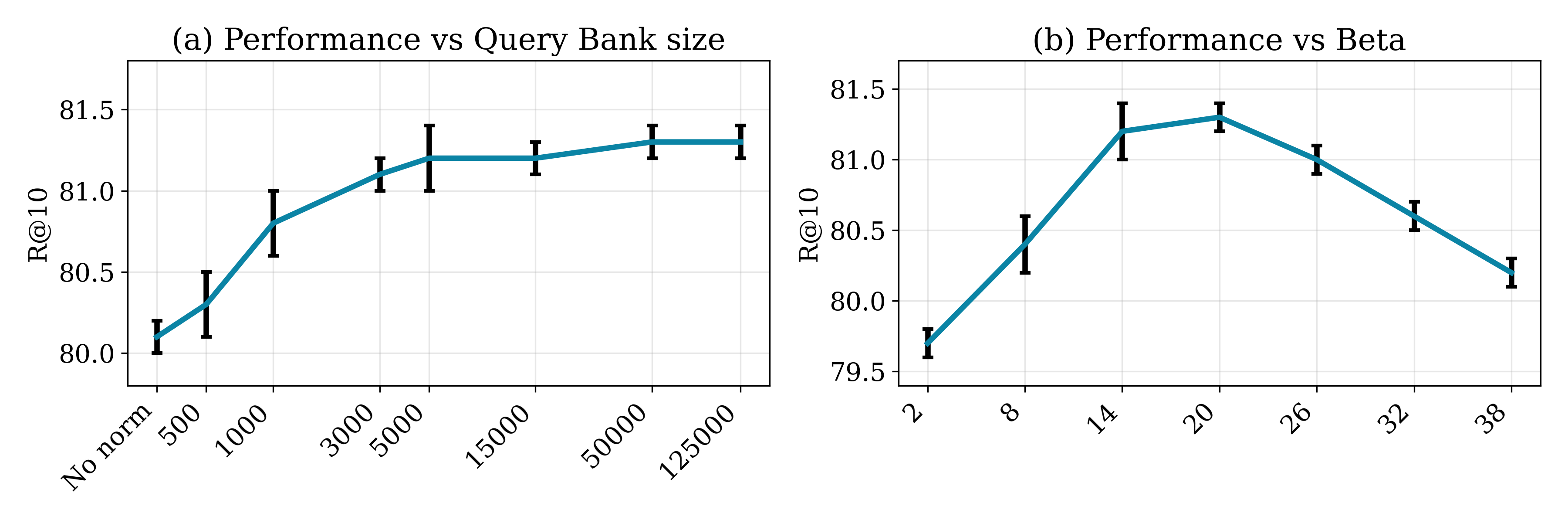}
    \end{subfigure}
    \vspace{-0.3cm}
    \caption{Retrieval results reported for a TT-CE+~\cite{croitoru2021teachtext} model on the MSR-VTT~\cite{xu2016msr} \texttt{validation} split in terms of R@1, R@5 and R@10.
    \textit{Left:} \textit{The influence of querybank size on retrieval performance.}
    We observe that performance grows steadily with increasing querybank size, but saturates.
    \textit{Right:} \textit{The influence of inverse temperature, $\beta$.} Performance varies smoothly with inverse temperature, peaking at a value of 20.}
    \mbox{}\vspace{-0.7cm} \\
    \label{fig:num-queries-and-hyperparam-r1}
\end{figure}

In the main paper,
to maintain conciseness we report ablation plots for the influence of querybank size
and inverse temperature  using the geometric mean of R1, R5 and R10.
For completeness, in this section we show results on each metric individually.
As seen in Fig.~\ref{fig:num-queries-and-hyperparam-r1},
the individual metrics reflect the trend shown for the geometric means,
aligning with the results shown in the main paper.

\begin{figure*}
    \centering
    \begin{subfigure}[b]{\textwidth}
        \centering
        \includegraphics[trim={0cm 0.5cm 0cm 0.2cm},clip,width=0.8\linewidth]{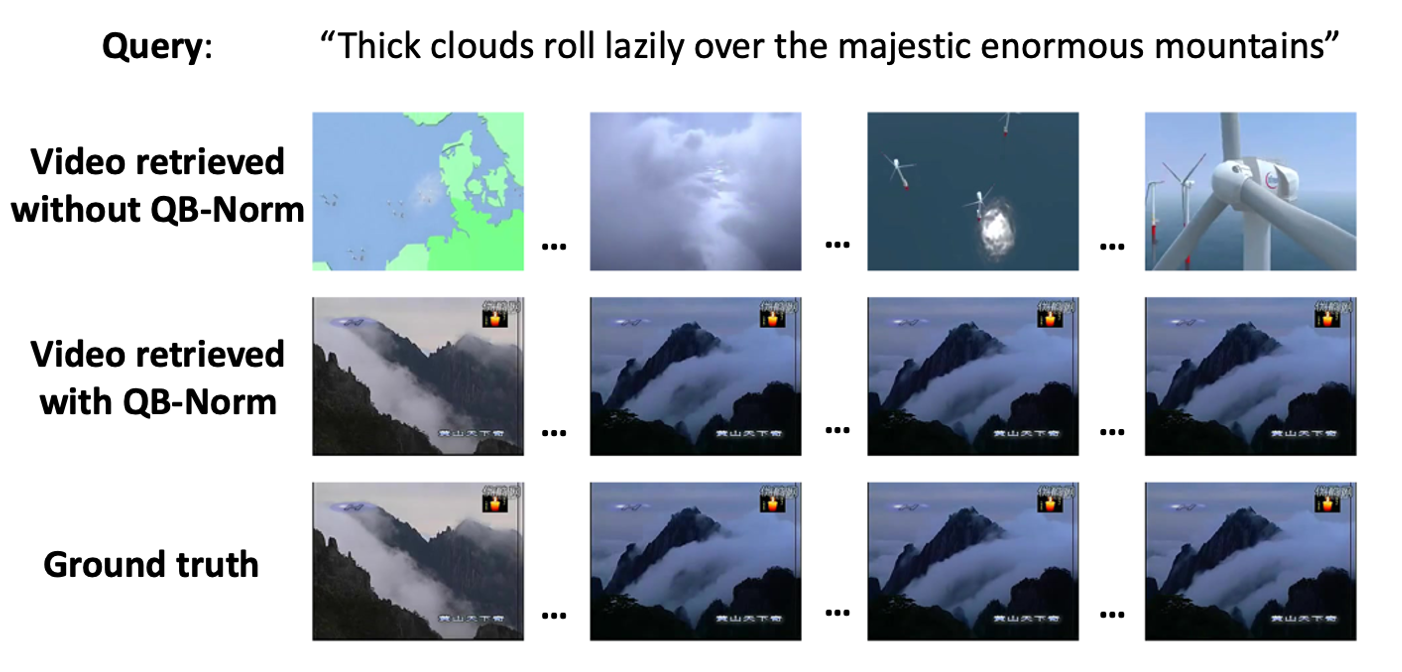}
        \caption{}
    \end{subfigure}
    \begin{subfigure}[b]{\textwidth}
        \centering
        \includegraphics[trim={0cm 0.5cm 0cm 0.2cm},clip,width=0.8\linewidth]{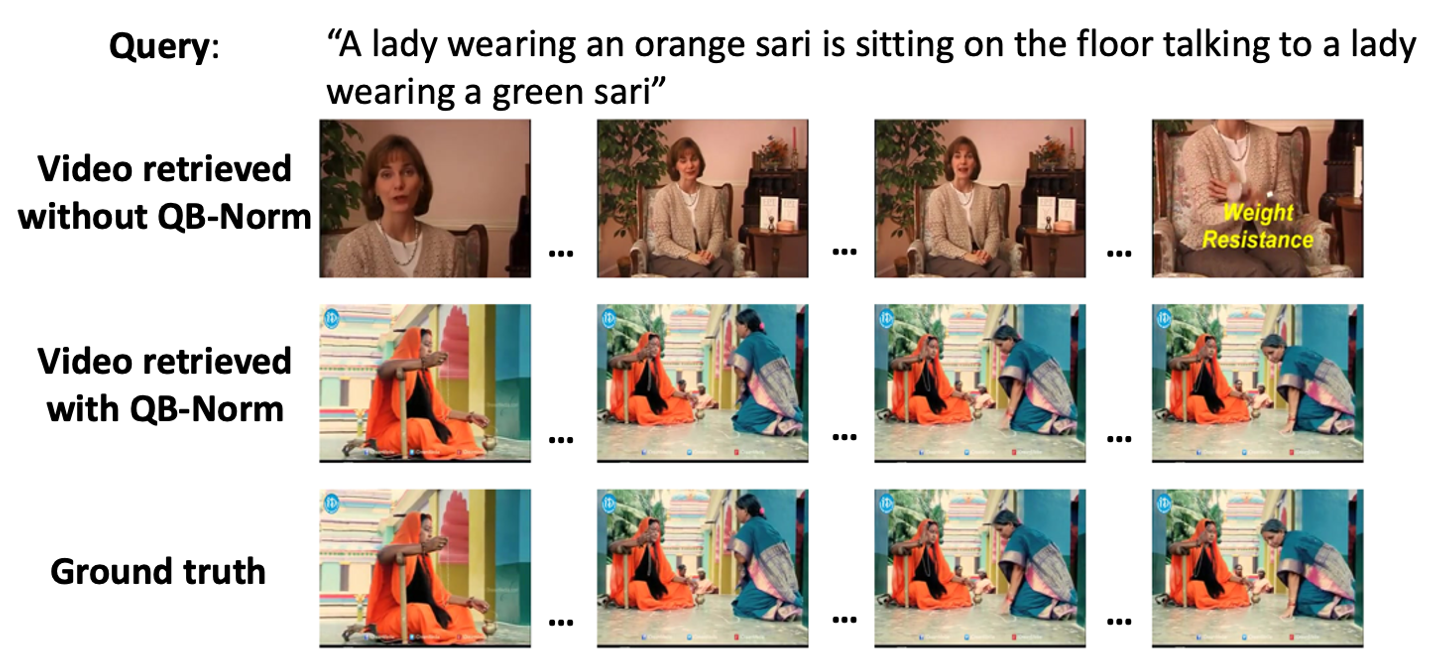}
        \caption{}
    \end{subfigure}
    \begin{subfigure}[b]{\textwidth}
        \centering
        \includegraphics[trim={0cm 0.5cm 0cm 0.2cm},clip,width=0.8\linewidth]{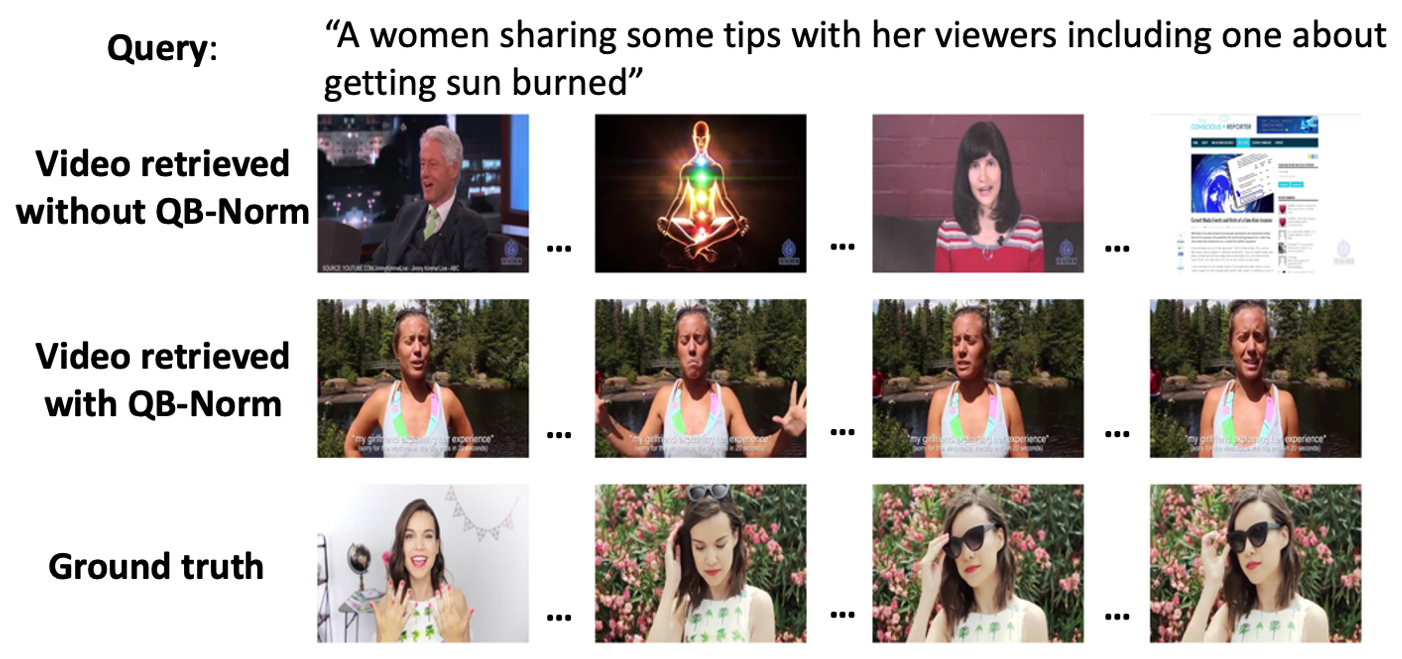}
        \caption{}
    \end{subfigure}
    \vspace{-0.3cm}
    \caption{\textbf{Qualitative results for the text video retrieval task.} 
    We show queries and frames from the retrieved videos.
    For the first two example queries,
    we observe that the use of \methodName leads to the retrieval of the
    correct target video.
    The third query represents a failure case in which the target video is not
    retrieved.
    However, we nevertheless observe qualitatively
    that for this example, the video retrieved
    with \methodName is more related to the query
    than the video retrieved without \methodName.
    }
    \mbox{}\vspace{-0.7cm} \\
    \label{fig:qualitative}
\end{figure*}

\section{Video and text embeddings (experts) description used for video retrieval}
\label{sec:experts}
For this work, we used the pretrained weights provided by TT-CE+ \cite{croitoru2021teachtext} and CE+ \cite{liu2019use} (\url{https://github.com/albanie/collaborative-experts}). These models use a set of pretrained experts. Below, we summarise how these experts were extracted.
\begin{itemize}
    \item Two action experts are used: \textit{Action(KN)} and \textit{Action(IG)}. \textit{Action(KN)} is a 1024-dimensional embedding produced by an I3D architecture trained on Kinetics~\cite{carreira2017quo}. The embeddings are extracted from frame clips at 25fps and center cropped to 224 pixels. For \textit{Action(IG)} the model is a 34-layer R(2+1)D~\cite{tran2018closer}, trained on IG-65m~\cite{ghadiyaram2019large}
    \item Two forms of object experts: \textit{Obj(IN)} and \textit{Obj(IG)}. For extracting \textit{Obj(IN)} a SENet-154~\cite{hu2019squeeze} model trained on ImageNet was used. For extracting \textit{Obj(IG)} a ResNext-101~\cite{xie2017aggregated} model trained on Instagram data with weakly labelled hashtags ~\cite{mahajan2018exploring} was used. Both of the embeddings are extracted at 25fps. 
    \item For producing an audio expert a VGGish model trained from audio classification on the YouTube-8m dataset~\cite{hershey2017} was used.
    \item For the scene expert a DenseNet-161~\cite{huang2017densely} pretrained on Places365~\cite{zhou2017places} was used. The scene embedding has a 2208 dimension.
    \item For the speech expert, the Google Cloud API (to transcribe the speech content) is used.
    \item For the text we use GPT2-xl~\cite{radford2019language} finetuned as provided by the authors. The size of the final pre-trained embedding is 1600.
\end{itemize}

For CLIP2Video~\cite{fang2021clip2video} we used the model as it is provided online \url{https://github.com/CryhanFang/CLIP2Video}. The model receives as input the raw frames and raw queries. For CLIP4Clip~\cite{luo2021clip4clip}, we use the online code \url{https://github.com/ArrowLuo/CLIP4Clip} and re-train the model for each dataset where we present results since weights are not available online.
For the other tasks we followed the instructions given on the official repositories. For MMT-Oscar~\cite{geigle2021retrieve} we used the pretrained weights and the features provided at \url{https://github.com/UKPLab/MMT-Retrieval}. For RDML~\cite{roth2020revisiting} we used the models provided at \url{https://github.com/Confusezius/Deep-Metric-Learning-Baselines}. For audio retrieval~\cite{oncescu2021audio} we used the pretrained weights and models provided at \url{https://github.com/oncescuandreea/audio-retrieval}.

\section{v2t performance metrics}
\label{sec:t2v}
\input{tables/msrvtt_v2t}
In Tab.~\ref{tab:msrvtt-v2t} we report metrics indicating the
performance of \methodName on the reverse task of video-text retrieval
(in which videos are used as queries to retrieve descriptions).
We apply \methodName with DIS normalisation using all videos from the training split
to construct the querybank.
We observe that \methodName yields a striking boost in performance.

\section{Qualitative results}
\label{sec:qualitative}
In Fig.~\ref{fig:qualitative} we provide some qualitative examples,
illustrating cases for which the \methodName model correctly retrieves
videos that are not retrieved without \methodName.
Examining failure cases, we found qualitative examples for which
the retrieval ranking produced with \methodName was more ``reasonable''
(as shown in the bottom set of Fig.~\ref{fig:qualitative}).
However, in line with prior work~\cite{radovanovic2010hubs} suggesting that
hubness is a property of the distribution
(rather than driven by individual samples),
we did not observe consistent, obvious qualitative trends
among the samples that were corrected, or remaining failure cases.
As an example, we observed gains for queries with both shorter
and longer, highly descriptive captions.

%% file: tables/topk-activations.tex
\begin{table}
\begin{center}
\resizebox{\linewidth}{!}{
\begin{tabular}{c|c|c|c|c|c}%
\hline%
\hline%
Querybank Source Data & Topk & $R@1\uparrow$ & $R@5\uparrow$  & $R@10\uparrow$ &  $MdR\downarrow$\\%
\hline%
\textbf{No querybank} & - &$14.9_{\pm0.1}$&$38.3_{\pm0.1}$&$51.5_{\pm0.1}$&$10.0_{\pm0.0}$\\%
\hline
\hline
\textit{In Domain} \\
\hline
\textbf{MSR-VTT}  & 1
&$17.0_{\pm0.1}$&$41.3_{\pm0.1}$&$54.1_{\pm0.1}$&$8.6_{\pm0.5}$\\%
\textbf{MSR-VTT}  & 2
&$17.1_{\pm0.1}$&$41.7_{\pm0.1}$&$54.5_{\pm0.1}$&$8.0_{\pm0.0}$\\%
\textbf{MSR-VTT}  & 3
&$17.1_{\pm0.1}$&$41.8_{\pm0.1}$&$54.6_{\pm0.1}$&$8.0_{\pm0.0}$\\%
\textbf{MSR-VTT}  & 5
&$17.1_{\pm0.1}$&$41.9_{\pm0.1}$&$54.7_{\pm0.1}$&$8.0_{\pm0.0}$\\%
\textbf{MSR-VTT}  & 10 &$17.1_{\pm0.1}$&$41.9_{\pm0.1}$&$54.7_{\pm0.1}$&$8.0_{\pm0.0}$\\%
\hline
\hline
\textit{Far Domain} \\
\hline
\textbf{LSMDC} & 1
&$14.9_{\pm0.1}$&$38.3_{\pm0.1}$&$51.2_{\pm0.1}$&$10.0_{\pm0.0}$\\%
\textbf{LSMDC} & 2
&$14.8_{\pm0.0}$&$38.0_{\pm0.0}$&$51.0_{\pm0.0}$&$10.0_{\pm0.0}$\\%
\textbf{LSMDC} & 3
&$14.7_{\pm0.0}$&$37.9_{\pm0.0}$&$50.9_{\pm0.0}$&$10.0_{\pm0.0}$\\%
\textbf{LSMDC} & 5
&$14.6_{\pm0.0}$&$37.8_{\pm0.0}$&$50.8_{\pm0.0}$&$10.0_{\pm0.0}$\\%
\textbf{LSMDC} & 10
&$14.5_{\pm0.0}$&$37.5_{\pm0.0}$&$50.4_{\pm0.0}$&$10.0_{\pm0.0}$\\%
\hline
\end{tabular}}
\end{center}
\vspace{\spacebefore{}}
\caption{\textbf{
The influence of the $k$ hyperparameter on DIS normalisation.} 
Performance is reported on MSR-VTT \texttt{full} split~\cite{xu2016msr},
while querybanks of 5,000 samples
are sampled from the training sets of different datasets.
We observe that for \textit{Far Domain} querybanks,
$k=1$ performs the best,
while retaining good performance for \textit{In Domain} querybanks.
\label{tab:topk-activations}}
\vspace{\spaceafter{}}
\end{table}

%% file: tables/querybank-train-vs-test-is.tex
\begin{table}
\begin{center}
\resizebox{\linewidth}{!}{
\begin{tabular}{c|c|c|c|c|c}%
\hline%
\hline%
Querybank Source&Size&$R@1\uparrow$&$R@5\uparrow$&$R@10\uparrow$&$MdR\downarrow$\\%
\hline%
\textbf{No querybank} & - &$14.9_{\pm0.1}$&$38.3_{\pm0.1}$&$51.5_{\pm0.1}$&$10.0_{\pm0.0}$\\%
\textbf{Training set} & 60k &$17.3_{\pm0.1}$&$42.1_{\pm0.2}$&$54.9_{\pm0.1}$&$8.0_{\pm0.0}$\\%

\textbf{Val set} & 10k
&$16.7_{\pm0.1}$&$41.2_{\pm0.1}$&$54.0_{\pm0.1}$&$8.7_{\pm0.5}$\\%

\textbf{Test set} & 60k &$17.5_{\pm0.0}$&$42.4_{\pm0.1}$&$55.1_{\pm0.0}$&$8.0_{\pm0.0}$\\%
\hline%
\end{tabular}}
\end{center}
\vspace{\spacebefore{}}
\caption{\textbf{Effective querybanks can be constructed
from the training set.}
Performance is reported on MSR-VTT
\texttt{full} split~\cite{xu2016msr} using IS normalisation.
We observe that a querybank of 60K samples from the training
set performs comparably to a test set querybank.
\label{tab:querybank-train-vs-test-is}}
\vspace{\spaceafter{}}
\end{table}

%% file: tables/supp-sota.tex
\begin{table}
\begin{center}
\resizebox{\linewidth}{!}{
\begin{tabular}{c|c|c|c|c}%
\hline%
\hline%
Model&$R@1\uparrow$&$R@5\uparrow$&$R@10\uparrow$&$MdR\downarrow$\\%
\hline%
Dual\cite{dong2019dual}&$7.7$&$22.0$&$31.8$&$32.0$\\%
HGR\cite{chen2020fine}&$9.2$&$26.2$&$36.5$&$24.0$\\%
MoEE\cite{miech2018learning}&$11.1_{\pm0.1}$&$30.7_{\pm0.1}$&$42.9_{\pm0.1}$&$15.0_{\pm0.0}$\\%
CE\cite{liu2019use}&$11.0_{\pm0.0}$&$30.8_{\pm0.1}$&$43.3_{\pm0.3}$&$15.0_{\pm0.0}$\\%
\hline%
CE+\cite{croitoru2021teachtext}&$14.4_{\pm0.1}$&$37.4_{\pm0.1}$&$50.2_{\pm0.1}$&$10.0_{\pm0.0}$\\%
\textbf{CE+ (+\methodName)}&$16.4_{\pm0.0}$&$40.3_{\pm0.1}$&$53.0_{\pm0.1}$&$9.0_{\pm0.0}$\\%
\hline
TT-CE+\cite{croitoru2021teachtext}&$14.9_{\pm0.1}$&$38.3_{\pm0.1}$&$51.5_{\pm0.1}$&$10.0_{\pm0.0}$\\%
\textbf{TT-CE+ (+\methodName)}&$17.3_{\pm0.0}$&$42.1_{\pm0.1}$&$54.9_{\pm0.1}$&$8.0_{\pm0.0}$\\%
\hline
CLIP4Clip\cite{luo2021clip4clip}$\ssymbol{3}$&$27.9$&$52.7$&$63.6$&$5.0$\\
\textbf{CLIP4Clip (+QB-Norm)}&$\mathbf{29.6}$&$\mathbf{54.5}$&$\mathbf{65.3}$&$\mathbf{4.0}$\\
\hline%
\end{tabular}}
\end{center}
\vspace{\spacebefore{}}
\caption{\textbf{MSR-VTT full split: comparison to state of the art.}\\$\ssymbol{3}$ denotes results obtained training using the official code.} \label{tab:msrvtt-final-sota}
\vspace{\spaceafter{}}
\end{table}

\begin{table}
\begin{center}
\resizebox{\linewidth}{!}{
\begin{tabular}{c|c|c|c|c}%
\hline%
\hline%
Model&$R@1\uparrow$&$R@5\uparrow$&$R@10\uparrow$&$MdR\downarrow$\\%
\hline%
MoEE\cite{miech2018learning}&$16.1_{\pm1.0}$&$41.2_{\pm1.6}$&$55.2_{\pm1.6}$&$8.3_{\pm0.5}$\\%
CE\cite{liu2019use}&$17.1_{\pm0.9}$&$41.9_{\pm0.2}$&$56.0_{\pm0.5}$&$8.0_{\pm0.0}$\\%
TT-CE&$21.0_{\pm0.6}$&$47.5_{\pm0.9}$&$61.9_{\pm0.5}$&$6.0_{\pm0.0}$\\%
Frozen\cite{bain2021frozen}&$31.0$&$59.8$&$72.4$&$3.0$\\%
CLIP4Clip~\cite{luo2021clip4clip}&$\mathbf{43.4}$&$70.2$&$80.6$&$\mathbf{2.0}$\\
\hline%
CE+\cite{croitoru2021teachtext}&$18.2_{\pm0.2}$&$43.9_{\pm0.9}$&$57.1_{\pm0.8}$&$7.9_{\pm0.1}$\\%
\textbf{CE+ (+\methodName)}&$20.7_{\pm0.6}$&$46.6_{\pm0.2}$&$59.8_{\pm0.2}$&$6.3_{\pm0.5}$\\%
\hline
TT-CE+\cite{croitoru2021teachtext}&$21.6_{\pm0.7}$&$48.6_{\pm0.4}$&$62.9_{\pm0.6}$&$6.0_{\pm0.0}$\\%
\textbf{TT-CE+ (+\methodName)}&$24.2_{\pm0.7}$&$50.8_{\pm0.7}$&$64.4_{\pm0.1}$&$5.3_{\pm0.5}$\\%
\hline
CLIP4Clip~\cite{luo2021clip4clip}$\ssymbol{3}$&$43.0$&$70.5$&$80.0$&$\mathbf{2.0}$\\%
\textbf{CLIP4Clip (+\methodName)}&$43.3$&$\mathbf{71.4}$&$\mathbf{80.8}$&$\mathbf{2.0}$\\%
\hline
\end{tabular}}
\end{center}
\vspace{\spacebefore{}}
\caption{\textbf{DiDeMo: Comparison to state of the art methods}.\\$\ssymbol{3}$ denotes results obtained training using the official code. 
\label{tab:didemo-final-sota}}
\vspace{\spaceafter{}}
\end{table}

\begin{table}
\begin{center}
\resizebox{\linewidth}{!}{
\begin{tabular}{c|c|c|c|c}%
\hline%
\hline%
Model&$R@1\uparrow$&$R@5\uparrow$&$R@10\uparrow$&$MdR\downarrow$\\%
\hline%
MoEE\cite{miech2018learning}&$12.1_{\pm0.7}$&$29.4_{\pm0.8}$&$37.7_{\pm0.2}$&$23.2_{\pm0.8}$\\%
CE\cite{liu2019use}&$12.4_{\pm0.7}$&$28.5_{\pm0.8}$&$37.9_{\pm0.6}$&$21.7_{\pm0.6}$\\%
MMT\cite{gabeur2020multi}&$13.2_{\pm0.4}$&$29.2_{\pm0.8}$&$38.8_{\pm0.9}$&$21.0_{\pm1.4}$\\%
Frozen\cite{bain2021frozen}&$15.0$&$30.8$&$39.8$&$20.0$\\%
CLIP4Clip~\cite{luo2021clip4clip}&$21.6$&$\mathbf{41.8}$&$\mathbf{49.8}$&$\mathbf{11.0}$\\
\hline%
CE+\cite{croitoru2021teachtext}&$14.9_{\pm0.6}$&$33.7_{\pm0.2}$&$44.1_{\pm0.6}$&$15.3_{\pm0.5}$\\%
\textbf{CE+ (\methodName)}&$16.4_{\pm0.8}$&$34.8_{\pm0.4}$&$44.9_{\pm0.9}$&$14.5_{\pm0.4}$\\%
\hline
TT-CE+\cite{croitoru2021teachtext}&$17.2_{\pm0.4}$&$36.5_{\pm0.6}$&$46.3_{\pm0.3}$&$13.7_{\pm0.5}$\\%
\textbf{TT-CE+ (\methodName)}&$17.8_{\pm0.4}$&$37.7_{\pm0.5}$&$47.6_{\pm0.6}$&$12.7_{\pm0.5}$\\%
\hline%
CLIP4Clip~\cite{luo2021clip4clip}$\ssymbol{3}$&$21.3$&$40.0$&$49.5$&$\mathbf{11.0}$\\%
\textbf{CLIP4Clip (+\methodName)}&$\mathbf{22.4}$&$40.1$&$49.5$&$\mathbf{11.0}$\\%
\hline%
\end{tabular}}
\end{center}
\vspace{\spacebefore{}}
\caption{\textbf{LSMDC: Comparison to state of the art methods}.\\$\ssymbol{3}$ denotes results obtained training using the official code.
\label{tab:lsmdc-final-sota}}
\vspace{\spaceafter{}}
\end{table}

\begin{table}
\begin{center}
\resizebox{\linewidth}{!}{
\begin{tabular}{c|c|c|c|c}%
\hline%
\hline%
Model&$R@1\uparrow$&$R@5\uparrow$&$R@50\uparrow$&$MdR\downarrow$\\%
\hline%
MoEE\cite{miech2018learning}&$19.7_{\pm0.3}$&$50.0_{\pm0.5}$&$92.0_{\pm0.2}$&$5.3_{\pm0.5}$\\%
CE\cite{liu2019use}&$19.9_{\pm0.3}$&$50.1_{\pm0.7}$&$92.2_{\pm0.6}$&$5.3_{\pm0.5}$\\%
HSE\cite{zhang2018cross}&$20.5$&$49.3$&$-$&$-$\\%
MMT\cite{gabeur2020multi}&$22.7_{\pm0.2}$&$54.2_{\pm1.0}$&$93.2_{\pm0.4}$&$5.0_{\pm0.0}$\\%
SSB\cite{patrick2020support}&$26.8$&$58.1$&$93.5$&$3.0$\\%
CLIP4Clip\cite{luo2021clip4clip}&$40.5$&$\mathbf{72.4}$&$\mathbf{98.1}$&$\mathbf{2.0}$\\
\hline
TT-CE+\cite{croitoru2021teachtext}&$23.5_{\pm0.2}$&$57.2_{\pm0.5}$&$96.1_{\pm0.1}$&$4.0_{\pm0.0}$\\%
\textbf{TT-CE+ (+\methodName)}&$27.0_{\pm0.2}$&$60.6_{\pm0.4}$&$96.8_{\pm0.0}$&$4.0_{\pm0.0}$\\%
\hline%
CLIP4Clip~\cite{luo2021clip4clip}$\ssymbol{3}$&$36.3$&$65.9$&$96.8$&$3.0$\\
\textbf{CLIP4Clip (+QB-Norm)}&$\mathbf{41.4}$&$71.4$&$97.6$&$\mathbf{2.0}$\\
\hline
\end{tabular}}
\end{center}
\vspace{\spacebefore{}}
\caption{\textbf{ActivityNet: Comparison to state of the art methods}.\\$\ssymbol{3}$ denotes results obtained training using the official code. \label{tab:activity-net-final-sota}}
\vspace{\spaceafter{}}
\end{table}

%% file: tables/cent.tex
\begin{table}
\begin{center}
\resizebox{\linewidth}{!}{
\begin{tabular}{c|c|c|c|c}%
\hline%
\hline%
Model&$R@1\uparrow$&$R@5\uparrow$&$R@10\uparrow$&$MdR\downarrow$\\%
\hline%
Baseline&$15.0$&$38.4$&$51.5$&$10.0$\\%
CENT~\cite{suzuki2013centering}&$14.4$ & $37.2$ & $50.2$ & $10.0$ \\%
DIS&$17.3$ & $42.1$ & $54.9$ & $8.0$ \\%
\hline%
\end{tabular}}
\end{center}
\vspace{\spacebefore{}}
\caption{\textbf{MSR-VTT full split} Comparison with CENT for a seed of TT-CE+\cite{croitoru2021teachtext} model. \label{tab:msrvtt-cent}}
\vspace{\spaceafter{}}
\end{table}

%% file: tables/msrvtt_v2t.tex
\begin{table}
\begin{center}
\resizebox{\linewidth}{!}{
\begin{tabular}{c|c|c|c|c|c}%
\hline%
\hline%
Model&Task&$R@1\uparrow$&$R@5\uparrow$&$R@10\uparrow$&$MdR\downarrow$\\%
\hline%
CE+\cite{croitoru2021teachtext}&v2t&$22.7_{\pm0.5}$&$52.6_{\pm0.6}$&$66.3_{\pm0.2}$&$5.0_{\pm0.0}$\\%
\textbf{CE+ (+\methodName)}&v2t&$28.6_{\pm0.4}$&$58.9_{\pm0.5}$&$71.4_{\pm0.5}$&$4.0_{\pm0.0}$\\%
\hline
TT-CE+\cite{croitoru2021teachtext}&v2t&$24.6_{\pm0.3}$&$54.1_{\pm0.3}$&$67.5_{\pm0.5}$&$4.7_{\pm0.5}$\\%
\textbf{TT-CE+ (+\methodName)}&v2t&$30.1_{\pm0.4}$&$61.4_{\pm0.4}$&$73.2_{\pm0.4}$&$3.0_{\pm0.0}$\\%
\hline%
\end{tabular}}
\end{center}
\vspace{\spacebefore{}}
\caption{\textbf{MSR-VTT full split: Comparison to state of the art - v2t task.}\label{tab:msrvtt-v2t}}
\vspace{\spaceafter{}}
\end{table}

%% file: 00-main.bbl
\begin{thebibliography}{100}\itemsep=-1pt

\bibitem{anne2017localizing}
Lisa Anne~Hendricks, Oliver Wang, Eli Shechtman, Josef Sivic, Trevor Darrell,
  and Bryan Russell.
\newblock Localizing moments in video with natural language.
\newblock In {\em Proceedings of the IEEE international conference on computer
  vision}, pages 5803--5812, 2017.

\bibitem{arandjelovic2017look}
Relja Arandjelovic and Andrew Zisserman.
\newblock Look, listen and learn.
\newblock In {\em Proceedings of the IEEE International Conference on Computer
  Vision}, pages 609--617, 2017.

\bibitem{Arandjelovi2018ObjectsTS}
Relja Arandjelovi{\'c} and Andrew Zisserman.
\newblock Objects that sound.
\newblock In {\em ECCV}, 2018.

\bibitem{aytar2008utilizing}
Yusuf Aytar, Mubarak Shah, and Jiebo Luo.
\newblock Utilizing semantic word similarity measures for video retrieval.
\newblock In {\em 2008 IEEE Conference on Computer Vision and Pattern
  Recognition}, pages 1--8. IEEE, 2008.

\bibitem{aytar2017see}
Yusuf Aytar, Carl Vondrick, and Antonio Torralba.
\newblock See, hear, and read: Deep aligned representations.
\newblock {\em arXiv preprint arXiv:1706.00932}, 2017.

\bibitem{bain2021frozen}
Max Bain, Arsha Nagrani, G{\"u}l Varol, and Andrew Zisserman.
\newblock Frozen in time: A joint video and image encoder for end-to-end
  retrieval.
\newblock 2021.

\bibitem{Page1961AdaptiveCP}
Richard~E. Bellman.
\newblock Adaptive control processes: A guided tour.
\newblock 1961.

\bibitem{Bentley1975MultidimensionalBS}
Jon~Louis Bentley.
\newblock Multidimensional binary search trees used for associative searching.
\newblock {\em Commun. ACM}, 18:509--517, 1975.

\bibitem{Berenzweig2007thesis}
Adam Berenzweig.
\newblock {\em Anchors and hubs in audio-based music similarity}.
\newblock PhD thesis, Columbia University, NY, USA, 2007.

\bibitem{blei2003modeling}
David~M Blei and Michael~I Jordan.
\newblock Modeling annotated data.
\newblock In {\em Proceedings of the 26th annual international ACM SIGIR
  conference on Research and development in informaion retrieval}, pages
  127--134, 2003.

\bibitem{brown2020smooth}
Andrew Brown, Weidi Xie, Vicky Kalogeiton, and Andrew Zisserman.
\newblock Smooth-ap: Smoothing the path towards large-scale image retrieval.
\newblock In {\em European Conference on Computer Vision}, pages 677--694.
  Springer, 2020.

\bibitem{bruce1986understanding}
Vicki Bruce and Andy Young.
\newblock Understanding face recognition.
\newblock {\em British journal of psychology}, 77(3):305--327, 1986.

\bibitem{bulat2021improving}
Adrian Bulat, Enrique S{\'a}nchez-Lozano, and Georgios Tzimiropoulos.
\newblock Improving memory banks for unsupervised learning with large
  mini-batch, consistency and hard negative mining.
\newblock In {\em ICASSP 2021-2021 IEEE International Conference on Acoustics,
  Speech and Signal Processing (ICASSP)}, pages 1695--1699. IEEE, 2021.

\bibitem{caba2015activitynet}
Fabian Caba~Heilbron, Victor Escorcia, Bernard Ghanem, and Juan Carlos~Niebles.
\newblock Activitynet: A large-scale video benchmark for human activity
  understanding.
\newblock In {\em Proceedings of the ieee conference on computer vision and
  pattern recognition}, pages 961--970, 2015.

\bibitem{carreira2017quo}
Joao Carreira and Andrew Zisserman.
\newblock Quo vadis, action recognition? a new model and the kinetics dataset.
\newblock In {\em proceedings of the IEEE Conference on Computer Vision and
  Pattern Recognition}, pages 6299--6308, 2017.

\bibitem{Chen2011CollectingHP}
David~L. Chen and William~B. Dolan.
\newblock Collecting highly parallel data for paraphrase evaluation.
\newblock In {\em ACL}, 2011.

\bibitem{chen2011collecting}
David~L Chen and William~B Dolan.
\newblock Collecting highly parallel data for paraphrase evaluation.
\newblock In {\em Proceedings of the 49th Annual Meeting of the Association for
  Computational Linguistics: Human Language Technologies-Volume 1}, pages
  190--200. Association for Computational Linguistics, 2011.

\bibitem{chenteam}
Shizhe Chen, Yida Zhao, and Qin Jin.
\newblock Team ruc ai.m3: Technical report in video pentathlon challenge 2020.

\bibitem{chen2020fine}
Shizhe Chen, Yida Zhao, Qin Jin, and Qi Wu.
\newblock Fine-grained video-text retrieval with hierarchical graph reasoning.
\newblock In {\em Proceedings of the IEEE/CVF Conference on Computer Vision and
  Pattern Recognition}, pages 10638--10647, 2020.

\bibitem{Chen2015MicrosoftCC}
Xinlei Chen, Hao Fang, Tsung-Yi Lin, Ramakrishna Vedantam, Saurabh Gupta, Piotr
  Doll{\'a}r, and C.~Lawrence Zitnick.
\newblock Microsoft coco captions: Data collection and evaluation server.
\newblock {\em ArXiv}, abs/1504.00325, 2015.

\bibitem{Cheraghian2019MitigatingTH}
Ali Cheraghian, Shafin Rahman, Dylan Campbell, and Lars Petersson.
\newblock Mitigating the hubness problem for zero-shot learning of 3d objects.
\newblock In {\em BMVC}, 2019.

\bibitem{chopra2005learning}
Sumit Chopra, Raia Hadsell, and Yann LeCun.
\newblock Learning a similarity metric discriminatively, with application to
  face verification.
\newblock In {\em 2005 IEEE Computer Society Conference on Computer Vision and
  Pattern Recognition (CVPR'05)}, volume~1, pages 539--546. IEEE, 2005.

\bibitem{Chum2011TotalRI}
Ondřej Chum, Andrej Mikul{\'i}k, Michal Perdoch, and Jiri Matas.
\newblock Total recall ii: Query expansion revisited.
\newblock {\em CVPR 2011}, pages 889--896, 2011.

\bibitem{Chum2007TotalRA}
Ondřej Chum, James Philbin, Josef Sivic, Michael Isard, and Andrew Zisserman.
\newblock Total recall: Automatic query expansion with a generative feature
  model for object retrieval.
\newblock {\em 2007 IEEE 11th International Conference on Computer Vision},
  pages 1--8, 2007.

\bibitem{conneau2018word}
Alexis Conneau, Guillaume Lample, Marc'Aurelio Ranzato, Ludovic Denoyer, and
  Herv{\'e} J{\'e}gou.
\newblock Word translation without parallel data.
\newblock In {\em International Conference on Learning Representations}, 2018.

\bibitem{cornia2020meshed}
Marcella Cornia, Matteo Stefanini, Lorenzo Baraldi, and Rita Cucchiara.
\newblock Meshed-memory transformer for image captioning.
\newblock In {\em Proceedings of the IEEE/CVF Conference on Computer Vision and
  Pattern Recognition}, pages 10578--10587, 2020.

\bibitem{croitoru2021teachtext}
Ioana Croitoru, Simion-Vlad Bogolin, Marius Leordeanu, Hailin Jin, Andrew
  Zisserman, Samuel Albanie, and Yang Liu.
\newblock Teachtext: Crossmodal generalized distillation for text-video
  retrieval.
\newblock In {\em Proceedings of the IEEE/CVF International Conference on
  Computer Vision}, pages 11583--11593, 2021.

\bibitem{dinu2014improving}
Georgiana Dinu, Angeliki Lazaridou, and Marco Baroni.
\newblock Improving zero-shot learning by mitigating the hubness problem.
\newblock {\em arXiv preprint arXiv:1412.6568}, 2014.

\bibitem{dong2016word2visualvec}
Jianfeng Dong, Xirong Li, and Cees~GM Snoek.
\newblock Word2visualvec: Image and video to sentence matching by visual
  feature prediction.
\newblock {\em arXiv preprint arXiv:1604.06838}, 2016.

\bibitem{dong2019dual}
Jianfeng Dong, Xirong Li, Chaoxi Xu, Shouling Ji, and Xun Wang.
\newblock Dual dense encoding for zero-example video retrieval.
\newblock In {\em Proceedings of the IEEE Conference on Computer Vision and
  Pattern Recognition}, 2019.

\bibitem{Dong2011EfficientKN}
Wei Dong, Moses Charikar, and K. Li.
\newblock Efficient k-nearest neighbor graph construction for generic
  similarity measures.
\newblock In {\em WWW}, 2011.

\bibitem{duygulu2002object}
Pinar Duygulu, Kobus Barnard, Joao~FG de Freitas, and David~A Forsyth.
\newblock Object recognition as machine translation: Learning a lexicon for a
  fixed image vocabulary.
\newblock In {\em European conference on computer vision}, pages 97--112.
  Springer, 2002.

\bibitem{faghri2017vse++}
Fartash Faghri, David~J Fleet, Jamie~Ryan Kiros, and Sanja Fidler.
\newblock Vse++: Improving visual-semantic embeddings with hard negatives.
\newblock {\em arXiv preprint arXiv:1707.05612}, 2017.

\bibitem{Faghri2018VSEIV}
Fartash Faghri, David~J. Fleet, Jamie~Ryan Kiros, and Sanja Fidler.
\newblock Vse++: Improving visual-semantic embeddings with hard negatives.
\newblock In {\em BMVC}, 2018.

\bibitem{fang2021clip2video}
Han Fang, Pengfei Xiong, Luhui Xu, and Yu Chen.
\newblock Clip2video: Mastering video-text retrieval via image clip.
\newblock {\em arXiv preprint arXiv:2106.11097}, 2021.

\bibitem{feldbauer2019comprehensive}
Roman Feldbauer and Arthur Flexer.
\newblock A comprehensive empirical comparison of hubness reduction in
  high-dimensional spaces.
\newblock {\em Knowledge and Information Systems}, 59(1):137--166, 2019.

\bibitem{Feldbauer2018FastAH}
Roman Feldbauer, Maximilian Leodolter, Claudia Plant, and Arthur Flexer.
\newblock Fast approximate hubness reduction for large high-dimensional data.
\newblock {\em 2018 IEEE International Conference on Big Knowledge (ICBK)},
  pages 358--367, 2018.

\bibitem{Feldbauer2020}
Roman Feldbauer, Thomas Rattei, and Arthur Flexer.
\newblock scikit-hubness: Hubness reduction and approximate neighbor search.
\newblock {\em Journal of Open Source Software}, 5(45):1957, 2020.

\bibitem{franccois2007concentration}
Damien Fran{\c{c}}ois, Vincent Wertz, and Michel Verleysen.
\newblock The concentration of fractional distances.
\newblock {\em IEEE Transactions on Knowledge and Data Engineering},
  19(7):873--886, 2007.

\bibitem{frome2013devise}
Andrea Frome, Greg Corrado, Jonathon Shlens, Samy Bengio, Jeffrey Dean,
  Marc’Aurelio Ranzato, and Tomas Mikolov.
\newblock Devise: A deep visual-semantic embedding model.
\newblock 2013.

\bibitem{gabeur2020multi}
Valentin Gabeur, Chen Sun, Karteek Alahari, and Cordelia Schmid.
\newblock Multi-modal transformer for video retrieval.
\newblock {\em European Conference on Computer Vision}, 2020.

\bibitem{geigle2021retrieve}
Gregor Geigle, Jonas Pfeiffer, Nils Reimers, Ivan Vuli{\'c}, and Iryna
  Gurevych.
\newblock Retrieve fast, rerank smart: Cooperative and joint approaches for
  improved cross-modal retrieval.
\newblock {\em arXiv preprint arXiv:2103.11920}, 2021.

\bibitem{ghadiyaram2019large}
Deepti Ghadiyaram, Du Tran, and Dhruv Mahajan.
\newblock Large-scale weakly-supervised pre-training for video action
  recognition.
\newblock In {\em Proceedings of the IEEE Conference on Computer Vision and
  Pattern Recognition}, pages 12046--12055, 2019.

\bibitem{Gong2011IterativeQA}
Yunchao Gong and Svetlana Lazebnik.
\newblock Iterative quantization: A procrustean approach to learning binary
  codes.
\newblock {\em CVPR 2011}, pages 817--824, 2011.

\bibitem{Graves2014NeuralTM}
Alex Graves, Greg Wayne, and Ivo Danihelka.
\newblock Neural turing machines.
\newblock {\em ArXiv}, abs/1410.5401, 2014.

\bibitem{greve2016evolving}
Rasmus~Boll Greve, Emil~Juul Jacobsen, and Sebastian Risi.
\newblock Evolving neural turing machines for reward-based learning.
\newblock In {\em Proceedings of the Genetic and Evolutionary Computation
  Conference 2016}, pages 117--124, 2016.

\bibitem{Hara2015LocalizedCR}
Kazuo Hara, Ikumi Suzuki, Masashi Shimbo, Kei Kobayashi, Kenji Fukumizu, and
  Milo{\vs} Radovanovi{\'c}.
\newblock Localized centering: Reducing hubness in large-sample data.
\newblock In {\em AAAI}, 2015.

\bibitem{He2020MomentumCF}
Kaiming He, Haoqi Fan, Yuxin Wu, Saining Xie, and Ross~B. Girshick.
\newblock Momentum contrast for unsupervised visual representation learning.
\newblock {\em 2020 IEEE/CVF Conference on Computer Vision and Pattern
  Recognition (CVPR)}, pages 9726--9735, 2020.

\bibitem{He2013KMeansHA}
Kaiming He, Fang Wen, and Jian Sun.
\newblock K-means hashing: An affinity-preserving quantization method for
  learning binary compact codes.
\newblock {\em 2013 IEEE Conference on Computer Vision and Pattern
  Recognition}, pages 2938--2945, 2013.

\bibitem{Hendricks2017LocalizingMI}
Lisa~Anne Hendricks, Oliver Wang, Eli Shechtman, Josef Sivic, Trevor Darrell,
  and Bryan~C. Russell.
\newblock Localizing moments in video with natural language.
\newblock {\em 2017 IEEE International Conference on Computer Vision (ICCV)},
  pages 5804--5813, 2017.

\bibitem{hershey2017}
Shawn Hershey, Sourish Chaudhuri, Daniel P.~W. Ellis, Jort~F. Gemmeke, Aren
  Jansen, Channing Moore, Manoj Plakal, Devin Platt, Rif~A. Saurous, Bryan
  Seybold, Malcolm Slaney, Ron Weiss, and Kevin Wilson.
\newblock Cnn architectures for large-scale audio classification.
\newblock In {\em International Conference on Acoustics, Speech and Signal
  Processing (ICASSP)}. 2017.

\bibitem{hu2019squeeze}
Jie Hu, Li Shen, Samuel Albanie, Gang Sun, and Enhua Wu.
\newblock Squeeze-and-excitation networks.
\newblock {\em IEEE transactions on pattern analysis and machine intelligence},
  2019.

\bibitem{huang2017densely}
Gao Huang, Zhuang Liu, Laurens Van Der~Maaten, and Kilian~Q Weinberger.
\newblock Densely connected convolutional networks.
\newblock In {\em Proceedings of the IEEE conference on computer vision and
  pattern recognition}, pages 4700--4708, 2017.

\bibitem{Indyk1998ApproximateNN}
Piotr Indyk and Rajeev Motwani.
\newblock Approximate nearest neighbors: towards removing the curse of
  dimensionality.
\newblock In {\em STOC '98}, 1998.

\bibitem{Jgou2008HammingEA}
Herv{\'e} J{\'e}gou, Matthijs Douze, and Cordelia Schmid.
\newblock Hamming embedding and weak geometric consistency for large scale
  image search.
\newblock In {\em ECCV}, 2008.

\bibitem{Jgou2011ProductQF}
Herv{\'e} J{\'e}gou, Matthijs Douze, and Cordelia Schmid.
\newblock Product quantization for nearest neighbor search.
\newblock {\em IEEE Transactions on Pattern Analysis and Machine Intelligence},
  33:117--128, 2011.

\bibitem{jegou2007contextual}
Herve Jegou, Hedi Harzallah, and Cordelia Schmid.
\newblock A contextual dissimilarity measure for accurate and efficient image
  search.
\newblock In {\em 2007 IEEE Conference on Computer Vision and Pattern
  Recognition}, pages 1--8. IEEE, 2007.

\bibitem{johnson2019billion}
Jeff Johnson, Matthijs Douze, and Herv{\'e} J{\'e}gou.
\newblock Billion-scale similarity search with gpus.
\newblock {\em IEEE Transactions on Big Data}, 2019.

\bibitem{Joulin2018LossIT}
Armand Joulin, Piotr Bojanowski, Tomas Mikolov, Herv{\'e} J{\'e}gou, and
  Edouard Grave.
\newblock Loss in translation: Learning bilingual word mapping with a retrieval
  criterion.
\newblock In {\em EMNLP}, 2018.

\bibitem{Kaiser2017LearningTR}
Lukasz Kaiser, Ofir Nachum, Aurko Roy, and Samy Bengio.
\newblock Learning to remember rare events.
\newblock {\em ArXiv}, abs/1703.03129, 2017.

\bibitem{Karpathy2017DeepVA}
Andrej Karpathy and Li Fei-Fei.
\newblock Deep visual-semantic alignments for generating image descriptions.
\newblock {\em IEEE Transactions on Pattern Analysis and Machine Intelligence},
  39:664--676, 2017.

\bibitem{kidron2005pixels}
Einat Kidron, Yoav~Y Schechner, and Michael Elad.
\newblock Pixels that sound.
\newblock In {\em 2005 IEEE Computer Society Conference on Computer Vision and
  Pattern Recognition (CVPR'05)}, volume~1, pages 88--95. IEEE, 2005.

\bibitem{Kim2019AbstractiveSO}
Byeongchang Kim, Hyunwoo Kim, and Gunhee Kim.
\newblock Abstractive summarization of reddit posts with multi-level memory
  networks.
\newblock {\em ArXiv}, abs/1811.00783, 2019.

\bibitem{Kim2019AudioCapsGC}
Chris~Dongjoo Kim, Byeongchang Kim, Hyunmin Lee, and Gunhee Kim.
\newblock Audiocaps: Generating captions for audios in the wild.
\newblock In {\em NAACL}, 2019.

\bibitem{kiros2014unifying}
Ryan Kiros, Ruslan Salakhutdinov, and Richard~S Zemel.
\newblock Unifying visual-semantic embeddings with multimodal neural language
  models.
\newblock {\em arXiv preprint arXiv:1411.2539}, 2014.

\bibitem{Krishna2017DenseCaptioningEI}
Ranjay Krishna, Kenji Hata, Frederic Ren, Li Fei-Fei, and Juan~Carlos Niebles.
\newblock Dense-captioning events in videos.
\newblock {\em 2017 IEEE International Conference on Computer Vision (ICCV)},
  pages 706--715, 2017.

\bibitem{Lee2018AMN}
Sangho Lee, Jinyoung Sung, Youngjae Yu, and Gunhee Kim.
\newblock A memory network approach for story-based temporal summarization of
  360° videos.
\newblock {\em 2018 IEEE/CVF Conference on Computer Vision and Pattern
  Recognition}, pages 1410--1419, 2018.

\bibitem{levi2021rethinking}
Elad Levi, Tete Xiao, Xiaolong Wang, and Trevor Darrell.
\newblock Rethinking preventing class-collapsing in metric learning with
  margin-based losses.
\newblock In {\em Proceedings of the IEEE/CVF International Conference on
  Computer Vision}, pages 10316--10325, 2021.

\bibitem{li2020oscar}
Xiujun Li, Xi Yin, Chunyuan Li, Xiaowei Hu, Pengchuan Zhang, Lei Zhang, Lijuan
  Wang, Houdong Hu, Li Dong, Furu Wei, Yejin Choi, and Jianfeng Gao.
\newblock Oscar: Object-semantics aligned pre-training for vision-language
  tasks.
\newblock {\em ECCV 2020}, 2020.

\bibitem{liu2019strong}
Fangyu Liu and Rongtian Ye.
\newblock A strong and robust baseline for text-image matching.
\newblock {\em arXiv preprint arXiv:1906.01205}, 2019.

\bibitem{Liu2020HALIT}
Fangyu Liu, Rongtian Ye, Xun Wang, and Shuaipeng Li.
\newblock Hal: Improved text-image matching by mitigating visual semantic hubs.
\newblock {\em ArXiv}, abs/1911.10097, 2020.

\bibitem{liu2018unsupervised}
Qun Liu and Supratik Mukhopadhyay.
\newblock Unsupervised learning using pretrained cnn and associative memory
  bank.
\newblock In {\em 2018 International Joint Conference on Neural Networks
  (IJCNN)}, pages 01--08. IEEE, 2018.

\bibitem{liu2019use}
Yang Liu, Samuel Albanie, Arsha Nagrani, and Andrew Zisserman.
\newblock Use what you have: Video retrieval using representations from
  collaborative experts.
\newblock {\em arXiv preprint arXiv:1907.13487}, 2019.

\bibitem{liu2021adaptive}
Yang Liu, Qingchao Chen, and Samuel Albanie.
\newblock Adaptive cross-modal prototypes for cross-domain visual-language
  retrieval.
\newblock In {\em Proceedings of the IEEE/CVF Conference on Computer Vision and
  Pattern Recognition}, pages 14954--14964, 2021.

\bibitem{low2013hubness}
Thomas Low, Christian Borgelt, Sebastian Stober, and Andreas N{\"u}rnberger.
\newblock The hubness phenomenon: Fact or artifact?
\newblock In {\em Towards Advanced Data Analysis by Combining Soft Computing
  and Statistics}, pages 267--278. Springer, 2013.

\bibitem{luo2021clip4clip}
Huaishao Luo, Lei Ji, Ming Zhong, Yang Chen, Wen Lei, Nan Duan, and Tianrui Li.
\newblock Clip4clip: An empirical study of clip for end to end video clip
  retrieval.
\newblock {\em arXiv preprint arXiv:2104.08860}, 2021.

\bibitem{mahajan2018exploring}
Dhruv Mahajan, Ross Girshick, Vignesh Ramanathan, Kaiming He, Manohar Paluri,
  Yixuan Li, Ashwin Bharambe, and Laurens van~der Maaten.
\newblock Exploring the limits of weakly supervised pretraining.
\newblock In {\em Proceedings of the European Conference on Computer Vision
  (ECCV)}, pages 181--196, 2018.

\bibitem{miech2021thinking}
Antoine Miech, Jean-Baptiste Alayrac, Ivan Laptev, Josef Sivic, and Andrew
  Zisserman.
\newblock Thinking fast and slow: Efficient text-to-visual retrieval with
  transformers.
\newblock In {\em Proceedings of the IEEE/CVF Conference on Computer Vision and
  Pattern Recognition}, pages 9826--9836, 2021.

\bibitem{miech2018learning}
Antoine Miech, Ivan Laptev, and Josef Sivic.
\newblock Learning a text-video embedding from incomplete and heterogeneous
  data.
\newblock {\em arXiv preprint arXiv:1804.02516}, 2018.

\bibitem{miech2019howto100m}
Antoine Miech, Dimitri Zhukov, Jean-Baptiste Alayrac, Makarand Tapaswi, Ivan
  Laptev, and Josef Sivic.
\newblock Howto100m: Learning a text-video embedding by watching hundred
  million narrated video clips.
\newblock In {\em Proceedings of the IEEE International Conference on Computer
  Vision}, pages 2630--2640, 2019.

\bibitem{mithun2018learning}
Niluthpol~Chowdhury Mithun, Juncheng Li, Florian Metze, and Amit~K
  Roy-Chowdhury.
\newblock Learning joint embedding with multimodal cues for cross-modal
  video-text retrieval.
\newblock In {\em Proceedings of the 2018 ACM on International Conference on
  Multimedia Retrieval}, pages 19--27, 2018.

\bibitem{Mithun2018WeblySJ}
Niluthpol~Chowdhury Mithun, Rameswar Panda, Evangelos~E. Papalexakis, and
  Amit~K. Roy-Chowdhury.
\newblock Webly supervised joint embedding for cross-modal image-text
  retrieval.
\newblock {\em Proceedings of the 26th ACM international conference on
  Multimedia}, 2018.

\bibitem{munro2021domain}
Jonathan Munro, Michael Wray, Diane Larlus, Gabriela Csurka, and Dima Damen.
\newblock Domain adaptation in multi-view embedding for cross-modal video
  retrieval.
\newblock {\em arXiv preprint arXiv:2110.12812}, 2021.

\bibitem{Nagrani2018LearnablePC}
Arsha Nagrani, Samuel Albanie, and Andrew Zisserman.
\newblock Learnable pins: Cross-modal embeddings for person identity.
\newblock In {\em ECCV}, 2018.

\bibitem{oncescu20queryd}
Andreea-Maria Oncescu, Joao~F. Henriques, Yang Liu, Andrew~Zisserman Zisserman,
  and Samuel Albanie.
\newblock Queryd: a video dataset with high-quality textual and audio
  narrations.
\newblock {\em arXiv preprint arXiv:2011.11071}, 2020.

\bibitem{oncescu2021audio}
Andreea-Maria Oncescu, A Koepke, Jo{\~a}o~F Henriques, Zeynep Akata, and Samuel
  Albanie.
\newblock Audio retrieval with natural language queries.
\newblock {\em Interspeech}, 2021.

\bibitem{owens2016ambient}
Andrew Owens, Jiajun Wu, Josh~H McDermott, William~T Freeman, and Antonio
  Torralba.
\newblock Ambient sound provides supervision for visual learning.
\newblock In {\em European conference on computer vision}, pages 801--816.
  Springer, 2016.

\bibitem{Park2017AttendTY}
Cesc~Chunseong Park, Byeongchang Kim, and Gunhee Kim.
\newblock Attend to you: Personalized image captioning with context sequence
  memory networks.
\newblock {\em 2017 IEEE Conference on Computer Vision and Pattern Recognition
  (CVPR)}, pages 6432--6440, 2017.

\bibitem{patrick2020support}
Mandela Patrick, Po-Yao Huang, Yuki Asano, Florian Metze, Alexander Hauptmann,
  Jo{\~a}o Henriques, and Andrea Vedaldi.
\newblock Support-set bottlenecks for video-text representation learning.
\newblock {\em arXiv preprint arXiv:2010.02824}, 2020.

\bibitem{Philbin2007ObjectRW}
James Philbin, Ondřej Chum, Michael Isard, Josef Sivic, and Andrew Zisserman.
\newblock Object retrieval with large vocabularies and fast spatial matching.
\newblock {\em 2007 IEEE Conference on Computer Vision and Pattern
  Recognition}, pages 1--8, 2007.

\bibitem{radford2021learning}
Alec Radford, Jong~Wook Kim, Chris Hallacy, Aditya Ramesh, Gabriel Goh,
  Sandhini Agarwal, Girish Sastry, Amanda Askell, Pamela Mishkin, Jack Clark,
  et~al.
\newblock Learning transferable visual models from natural language
  supervision.
\newblock {\em arXiv preprint arXiv:2103.00020}, 2021.

\bibitem{radford2019language}
Alec Radford, Jeffrey Wu, Rewon Child, David Luan, Dario Amodei, and Ilya
  Sutskever.
\newblock Language models are unsupervised multitask learners.
\newblock {\em OpenAI Blog}, 1(8):9, 2019.

\bibitem{radovanovic2010hubs}
Milo{\v{s}} Radovanovi{\'c}, Alexandros Nanopoulos, and Mirjana Ivanovi{\'c}.
\newblock Hubs in space: Popular nearest neighbors in high-dimensional data.
\newblock {\em Journal of Machine Learning Research}, 11(Sep):2487--2531, 2010.

\bibitem{rasiwasia2010new}
Nikhil Rasiwasia, Jose Costa~Pereira, Emanuele Coviello, Gabriel Doyle, Gert~RG
  Lanckriet, Roger Levy, and Nuno Vasconcelos.
\newblock A new approach to cross-modal multimedia retrieval.
\newblock In {\em Proceedings of the 18th ACM international conference on
  Multimedia}, pages 251--260, 2010.

\bibitem{Rohrbach2015ADF}
Anna Rohrbach, Marcus Rohrbach, Niket Tandon, and Bernt Schiele.
\newblock A dataset for movie description.
\newblock {\em 2015 IEEE Conference on Computer Vision and Pattern Recognition
  (CVPR)}, pages 3202--3212, 2015.

\bibitem{rohrbach2017movie}
Anna Rohrbach, Atousa Torabi, Marcus Rohrbach, Niket Tandon, Christopher Pal,
  Hugo Larochelle, Aaron Courville, and Bernt Schiele.
\newblock Movie description.
\newblock {\em International Journal of Computer Vision}, 123(1):94--120, 2017.

\bibitem{roth2020revisiting}
Karsten Roth, Timo Milbich, Samarth Sinha, Prateek Gupta, Bjorn Ommer, and
  Joseph~Paul Cohen.
\newblock Revisiting training strategies and generalization performance in deep
  metric learning.
\newblock In {\em International Conference on Machine Learning}, pages
  8242--8252. PMLR, 2020.

\bibitem{santoro2016meta}
Adam Santoro, Sergey Bartunov, Matthew Botvinick, Daan Wierstra, and Timothy
  Lillicrap.
\newblock Meta-learning with memory-augmented neural networks.
\newblock In {\em International conference on machine learning}, pages
  1842--1850. PMLR, 2016.

\bibitem{schnitzer2012local}
Dominik Schnitzer, Arthur Flexer, Markus Schedl, and Gerhard Widmer.
\newblock Local and global scaling reduce hubs in space.
\newblock {\em Journal of Machine Learning Research}, 13(10), 2012.

\bibitem{shigeto2015ridge}
Yutaro Shigeto, Ikumi Suzuki, Kazuo Hara, Masashi Shimbo, and Yuji Matsumoto.
\newblock Ridge regression, hubness, and zero-shot learning.
\newblock In {\em Joint European conference on machine learning and knowledge
  discovery in databases}, pages 135--151. Springer, 2015.

\bibitem{slaney2002semantic}
Malcolm Slaney.
\newblock Semantic-audio retrieval.
\newblock In {\em 2002 IEEE International Conference on Acoustics, Speech, and
  Signal Processing}, volume~4, pages IV--4108. IEEE, 2002.

\bibitem{smith2017offline}
Samuel~L Smith, David~HP Turban, Steven Hamblin, and Nils~Y Hammerla.
\newblock Offline bilingual word vectors, orthogonal transformations and the
  inverted softmax.
\newblock {\em arXiv preprint arXiv:1702.03859}, 2017.

\bibitem{socher2010connecting}
Richard Socher and Li Fei-Fei.
\newblock Connecting modalities: Semi-supervised segmentation and annotation of
  images using unaligned text corpora.
\newblock In {\em 2010 IEEE Computer Society Conference on Computer Vision and
  Pattern Recognition}, pages 966--973. IEEE, 2010.

\bibitem{Socher2014GroundedCS}
Richard Socher, Andrej Karpathy, Quoc~V. Le, Christopher~D. Manning, and A. Ng.
\newblock Grounded compositional semantics for finding and describing images
  with sentences.
\newblock {\em Transactions of the Association for Computational Linguistics},
  2:207--218, 2014.

\bibitem{Song2016DeepML}
Hyun~Oh Song, Yu Xiang, Stefanie Jegelka, and Silvio Savarese.
\newblock Deep metric learning via lifted structured feature embedding.
\newblock {\em 2016 IEEE Conference on Computer Vision and Pattern Recognition
  (CVPR)}, pages 4004--4012, 2016.

\bibitem{Suzuki2012InvestigatingTE}
Ikumi Suzuki, Kazuo Hara, Masashi Shimbo, Yuji Matsumoto, and Marco Saerens.
\newblock Investigating the effectiveness of laplacian-based kernels in hub
  reduction.
\newblock In {\em AAAI}, 2012.

\bibitem{suzuki2013centering}
Ikumi Suzuki, Kazuo Hara, Masashi Shimbo, Marco Saerens, and Kenji Fukumizu.
\newblock Centering similarity measures to reduce hubs.
\newblock In {\em Proceedings of the 2013 conference on empirical methods in
  natural language processing}, pages 613--623, 2013.

\bibitem{thomee2016yfcc100m}
Bart Thomee, David~A Shamma, Gerald Friedland, Benjamin Elizalde, Karl Ni,
  Douglas Poland, Damian Borth, and Li-Jia Li.
\newblock Yfcc100m: The new data in multimedia research.
\newblock {\em Communications of the ACM}, 59(2):64--73, 2016.

\bibitem{tran2018closer}
Du Tran, Heng Wang, Lorenzo Torresani, Jamie Ray, Yann LeCun, and Manohar
  Paluri.
\newblock A closer look at spatiotemporal convolutions for action recognition.
\newblock In {\em Proceedings of the IEEE conference on Computer Vision and
  Pattern Recognition}, pages 6450--6459, 2018.

\bibitem{venugopalan2015sequence}
Subhashini Venugopalan, Marcus Rohrbach, Jeffrey Donahue, Raymond Mooney,
  Trevor Darrell, and Kate Saenko.
\newblock Sequence to sequence-video to text.
\newblock In {\em Proceedings of the IEEE international conference on computer
  vision}, pages 4534--4542, 2015.

\bibitem{Wah2011TheCB}
Catherine Wah, Steve Branson, Peter Welinder, Pietro Perona, and Serge~J.
  Belongie.
\newblock The caltech-ucsd birds-200-2011 dataset.
\newblock 2011.

\bibitem{Wang2016LearningDS}
Liwei Wang, Yin Li, and Svetlana Lazebnik.
\newblock Learning deep structure-preserving image-text embeddings.
\newblock {\em 2016 IEEE Conference on Computer Vision and Pattern Recognition
  (CVPR)}, pages 5005--5013, 2016.

\bibitem{wang2019multi}
Xun Wang, Xintong Han, Weilin Huang, Dengke Dong, and Matthew~R Scott.
\newblock Multi-similarity loss with general pair weighting for deep metric
  learning.
\newblock In {\em Proceedings of the IEEE/CVF Conference on Computer Vision and
  Pattern Recognition}, pages 5022--5030, 2019.

\bibitem{wang2019vatex}
Xin Wang, Jiawei Wu, Junkun Chen, Lei Li, Yuan-Fang Wang, and William~Yang
  Wang.
\newblock Vatex: A large-scale, high-quality multilingual dataset for
  video-and-language research.
\newblock In {\em Proceedings of the IEEE International Conference on Computer
  Vision}, pages 4581--4591, 2019.

\bibitem{wang2020cross}
Xun Wang, Haozhi Zhang, Weilin Huang, and Matthew~R Scott.
\newblock Cross-batch memory for embedding learning.
\newblock In {\em Proceedings of the IEEE/CVF Conference on Computer Vision and
  Pattern Recognition}, pages 6388--6397, 2020.

\bibitem{Wang2019VaTeXAL}
Xin~Eric Wang, Jiawei Wu, Junkun Chen, Lei Li, Yuan fang Wang, and William~Yang
  Wang.
\newblock Vatex: A large-scale, high-quality multilingual dataset for
  video-and-language research.
\newblock {\em 2019 IEEE/CVF International Conference on Computer Vision
  (ICCV)}, pages 4580--4590, 2019.

\bibitem{weston2011wsabie}
Jason Weston, Samy Bengio, and Nicolas Usunier.
\newblock Wsabie: Scaling up to large vocabulary image annotation.
\newblock In {\em Twenty-Second International Joint Conference on Artificial
  Intelligence}, 2011.

\bibitem{wray2019fine}
Michael Wray, Diane Larlus, Gabriela Csurka, and Dima Damen.
\newblock Fine-grained action retrieval through multiple parts-of-speech
  embeddings.
\newblock In {\em Proceedings of the IEEE International Conference on Computer
  Vision}, pages 450--459, 2019.

\bibitem{xie2017aggregated}
Saining Xie, Ross Girshick, Piotr Doll{\'a}r, Zhuowen Tu, and Kaiming He.
\newblock Aggregated residual transformations for deep neural networks.
\newblock In {\em Proceedings of the IEEE conference on computer vision and
  pattern recognition}, pages 1492--1500, 2017.

\bibitem{xu2020interactive}
Chunpu Xu, Yu Li, Chengming Li, Xiang Ao, Min Yang, and Jinwen Tian.
\newblock Interactive key-value memory-augmented attention for image paragraph
  captioning.
\newblock In {\em Proceedings of the 28th International Conference on
  Computational Linguistics}, pages 3132--3142, 2020.

\bibitem{xu2016msr}
Jun Xu, Tao Mei, Ting Yao, and Yong Rui.
\newblock Msr-vtt: A large video description dataset for bridging video and
  language.
\newblock In {\em Proceedings of the IEEE conference on computer vision and
  pattern recognition}, pages 5288--5296, 2016.

\bibitem{Xu2015JointlyMD}
Ran Xu, Caiming Xiong, Wei Chen, and Jason~J. Corso.
\newblock Jointly modeling deep video and compositional text to bridge vision
  and language in a unified framework.
\newblock In {\em AAAI}, 2015.

\bibitem{xu2015jointly}
Ran Xu, Caiming Xiong, Wei Chen, and Jason~J Corso.
\newblock Jointly modeling deep video and compositional text to bridge vision
  and language in a unified framework.
\newblock In {\em Twenty-Ninth AAAI Conference on Artificial Intelligence},
  2015.

\bibitem{Yoo2019ColoringWL}
Seungjoo Yoo, Hyojin Bahng, Sunghyo Chung, Junsoo Lee, Jaehyuk Chang, and
  Jaegul Choo.
\newblock Coloring with limited data: Few-shot colorization via memory
  augmented networks.
\newblock {\em 2019 IEEE/CVF Conference on Computer Vision and Pattern
  Recognition (CVPR)}, pages 11275--11284, 2019.

\bibitem{yu2018joint}
Youngjae Yu, Jongseok Kim, and Gunhee Kim.
\newblock A joint sequence fusion model for video question answering and
  retrieval.
\newblock In {\em Proceedings of the European Conference on Computer Vision
  (ECCV)}, pages 471--487, 2018.

\bibitem{zaremba2015reinforcement}
Wojciech Zaremba and Ilya Sutskever.
\newblock Reinforcement learning neural turing machines-revised.
\newblock {\em arXiv preprint arXiv:1505.00521}, 2015.

\bibitem{ZelnikManor2004SelfTuningSC}
Lihi Zelnik-Manor and Pietro Perona.
\newblock Self-tuning spectral clustering.
\newblock In {\em NIPS}, 2004.

\bibitem{zhang2018cross}
Bowen Zhang, Hexiang Hu, and Fei Sha.
\newblock Cross-modal and hierarchical modeling of video and text.
\newblock In {\em Proceedings of the European Conference on Computer Vision
  (ECCV)}, pages 374--390, 2018.

\bibitem{zhang2017learning}
Li Zhang, Tao Xiang, and Shaogang Gong.
\newblock Learning a deep embedding model for zero-shot learning.
\newblock In {\em Proceedings of the IEEE conference on computer vision and
  pattern recognition}, pages 2021--2030, 2017.

\bibitem{zhang2021vinvl}
Pengchuan Zhang, Xiujun Li, Xiaowei Hu, Jianwei Yang, Lei Zhang, Lijuan Wang,
  Yejin Choi, and Jianfeng Gao.
\newblock Vinvl: Making visual representations matter in vision-language
  models.
\newblock {\em CVPR 2021}, 2021.

\bibitem{zhao2018sound}
Hang Zhao, Chuang Gan, Andrew Rouditchenko, Carl Vondrick, Josh McDermott, and
  Antonio Torralba.
\newblock The sound of pixels.
\newblock In {\em Proceedings of the European conference on computer vision
  (ECCV)}, pages 570--586, 2018.

\bibitem{Zhao2020RUC}
Yida Zhao, Yuqing Song, Shizhe Chen, and Qin Jin.
\newblock Ruc\_aim3 at trecvid 2020: Ad-hoc video search \& video to text
  description.
\newblock In {\em TRECVID}, 2020.

\bibitem{zhou2017places}
Bolei Zhou, Agata Lapedriza, Aditya Khosla, Aude Oliva, and Antonio Torralba.
\newblock Places: A 10 million image database for scene recognition.
\newblock {\em IEEE Transactions on Pattern Analysis and Machine Intelligence},
  2017.

\end{thebibliography}
